\documentclass[letterpaper]{article} 
\usepackage{aaai25}  
\usepackage{times}  
\usepackage{helvet}  
\usepackage{courier}  
\usepackage[hyphens]{url}  
\usepackage{graphicx} 
\usepackage{natbib}  
\usepackage{caption} 
\usepackage{tabularx}
\usepackage{algorithm}
\usepackage{algorithmic}
\newcommand{\dataset}{VTG-IT} 
\newcommand{\alg}{VTG-LLM}
\usepackage{newfloat}
\usepackage{listings}
\usepackage{booktabs}
\usepackage{multirow}
\usepackage{transparent}
\usepackage{amsmath,amsthm,amssymb}

\providecommand{\R}{\mathbb{R}} 

\providecommand{\cT}{\mathcal{T}}

\renewcommand{\ss}{\mathbf{s}}

\providecommand{\zz}{\mathbf{z}}

\providecommand{\mW}{\mathbf{W}}

\DeclareMathOperator*{\argmin}{arg\,min}
\DeclareMathOperator*{\argmax}{arg\,max}

\usepackage{newfloat}
\usepackage{listings}
\DeclareCaptionStyle{ruled}{labelfont=normalfont,labelsep=colon,strut=off} 
\floatstyle{ruled}
\newfloat{listing}{tb}{lst}{}
\floatname{listing}{Listing}
\title{VTG-LLM: Integrating Timestamp Knowledge into Video LLMs for Enhanced Video Temporal Grounding}
\author{
    Yongxin Guo\textsuperscript{\rm 1}\thanks{ This work was done when Yongxin Guo was an intern at Tencent PCG.},
    Jingyu Liu\textsuperscript{\rm 2},
    Mingda Li\textsuperscript{\rm 2},
    Dingxin Cheng\textsuperscript{\rm 3},
    Xiaoying Tang\textsuperscript{\rm 1,\rm 4,\rm 5}\thanks{Corresponding authors.},
    Dianbo Sui\textsuperscript{\rm 6},
    Qingbin Liu\textsuperscript{\rm 2},
    Xi Chen\textsuperscript{\rm 2}\textsuperscript{\textdagger},
    Kevin Zhao\textsuperscript{\rm 2}
}
\affiliations{
    \textsuperscript{\rm 1} School of Science and Engineering, The Chinese University of Hong Kong, Shenzhen, Guangdong, 518172, P.R. China\\
    \textsuperscript{\rm 2} Tencent PCG \\
    \textsuperscript{\rm 3} Shandong University \\
    \textsuperscript{\rm 4} The Shenzhen Future Network of Intelligence Institute, CUHK-Shenzhen, 518172, P.R. China \\
    \textsuperscript{\rm 5} The Guangdong Provincial Key Laboratory of Future Networks of Intelligence, CUHK-Shenzhen, 518172, P.R. China \\
    \textsuperscript{\rm 6} Harbin Institute of Technology

    yongxinguo@link.cuhk.edu.cn,
    tangxiaoying@cuhk.edu.cn,
    jasonxchen@tencent.com
}

\usepackage{bibentry}

\begin{document}

\maketitle

\begin{abstract}
    Video Temporal Grounding (VTG) strives to accurately pinpoint event timestamps in a specific video using linguistic queries, significantly impacting downstream tasks like video browsing and editing. Unlike traditional task-specific models, Video Large Language Models (video LLMs) can handle multiple tasks concurrently in a zero-shot manner. Consequently, exploring the application of video LLMs for VTG tasks has become a burgeoning research area.
    However, despite considerable advancements in video content understanding, video LLMs often struggle to accurately pinpoint timestamps within videos, limiting their effectiveness in VTG tasks. To address this, we introduce VTG-LLM, a model designed to enhance video LLMs' timestamp localization abilities. Our approach includes: (1) effectively integrating timestamp knowledge into visual tokens; (2) incorporating absolute-time tokens to manage timestamp knowledge without concept shifts; and (3) introducing a lightweight, high-performance, slot-based token compression technique designed to accommodate the demands of a large number of frames to be sampled for VTG tasks.
    Additionally, we present VTG-IT-120K, a collection of publicly available VTG datasets that we have re-annotated to improve upon low-quality annotations.
    Our comprehensive experiments demonstrate the superior performance of VTG-LLM in comparison to other video LLM methods across a variety of VTG tasks. 
    \end{abstract}

    \begin{links}
        \link{Code}{https://github.com/gyxxyg/VTG-LLM}
    \end{links}

%
    
\section{Introduction}


\begin{figure*}[!t]
    \centering
    \includegraphics[width=1.\textwidth]{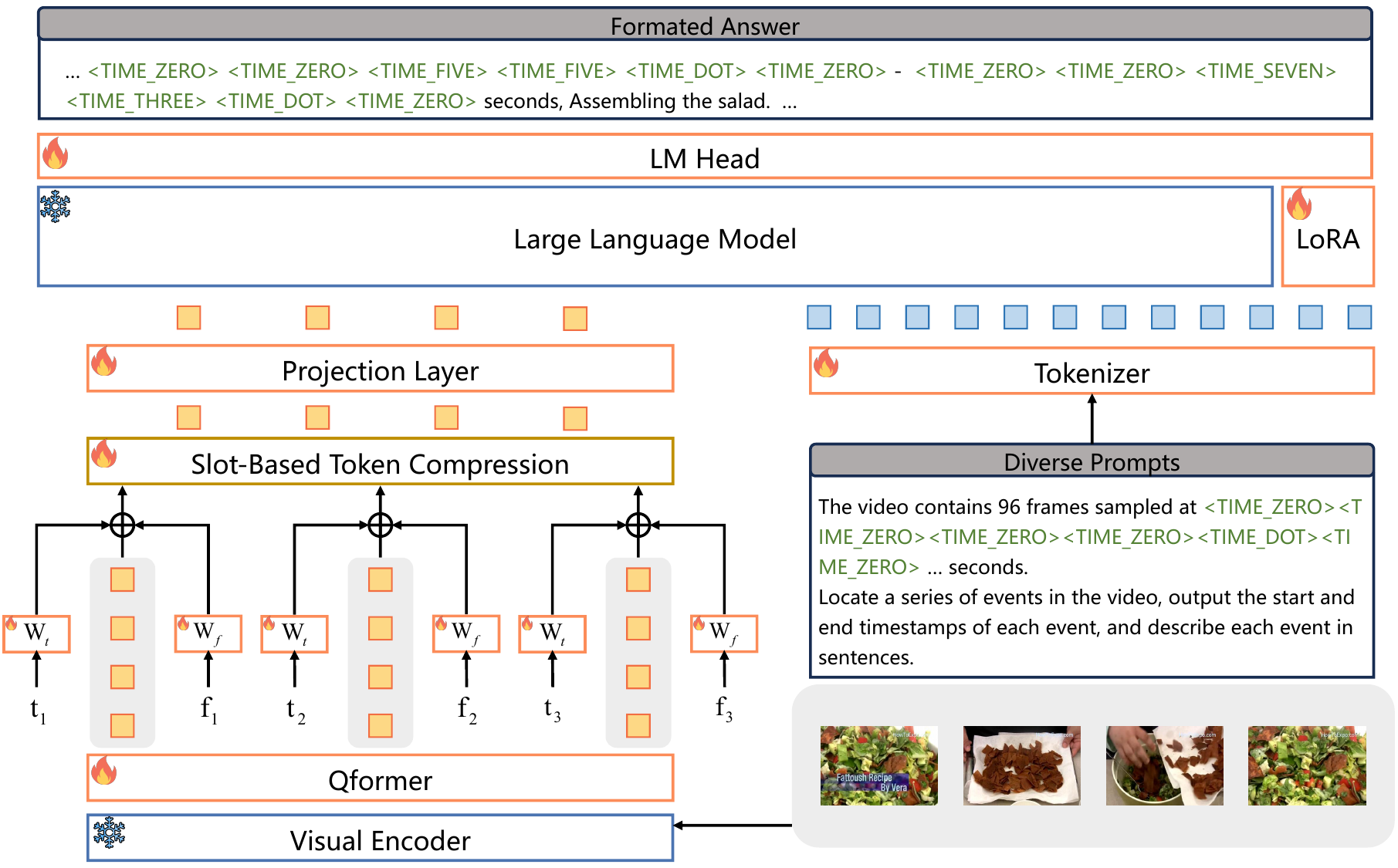}
    \caption{Overview of the VTG-LLM model.\looseness=-1}
    \label{fig:overview}
\end{figure*}

Video Temporal Grounding (VTG) is a crucial component of video understanding. It requires models to accurately pinpoint event timestamps within a video based on the given query. 
This task is essential for subsequent operations such as video browsing and editing.
In this paper, we adopt the categorization described by \citet{lin2023univtg}, and divide the VTG task into four primary sub-tasks.
For instance, moment retrieval and dense video captioning tasks~\citep{caba2015activitynet,zhou2018towards} require models to generate one or multiple timestamp intervals, each accompanied by captions. The video highlight detection task~\citep{lei2021detecting} necessitates models to produce a prominent score curve, while the video summarization task~\citep{song2015tvsum} demands that models output a series of timestamps, each associated with their corresponding video frames.

Despite significant efforts by traditional task-specific models designed for various VTG tasks~\citep{yang2023vid2seq, luo2023towards, lei2021detecting}, these methods are limited in their ability to (1) handle multiple VTG tasks simultaneously and (2) provide zero-shot capabilities on VTG tasks, which are crucial for real-world applications. 
As a remedy, recent research has begun exploring the use of video LLMs~\citep{lin2023video,zhang2023video,li2023videochat} as generalists~\citep{ren2023timechat, huang2023vtimellm} for addressing VTG tasks due to their capacity to handle various tasks concurrently in a zero-shot manner. However, several persistent issues hinder the effectiveness of current video LLMs in understanding timestamp knowledge, which in turn affects their performance on VTG tasks:

\begin{itemize}
    \item Visual inputs should contain sufficient and accurate timestamp information to help models understand when the visual content occurs in the videos.
    \item Concept shifts occur when varying input data produce identical output targets, potentially obscuring decision boundaries~\citep{moreno2012unifying}. This issue arises when using shared token embedding and classification heads for all digit-related knowledge. For example, the number '20' can appear in both a counting context, such as 'There are 20 people,' and a temporal context like 'From 20-30 seconds.' Although these digits have different meanings in these situations, they are forced to share the same decision boundary, making classification more challenging.
    \item VTG tasks necessitate sampling more frames compared to other video tasks, i.e., VQA tasks typically require only 8 or 16 frames~\citep{lin2023video,cheng2024videollama}. However, models cannot make reliable predictions on visual content that is not included in the sampled frames fed into language models. Therefore, due to the limited context length of LLMs, it is essential to develop effective token compression methods that enable the sampling of more frames.
\end{itemize}

To address these issues, in this paper, we present a novel video LLM model, VTG-LLM, comprising three components that efficiently integrate timestamp knowledge into video LLMs, thereby enhancing their performance on VTG tasks:
(1) we introduce a sequence-time embedding method that integrates accurate timestamp data into visual tokens;
(2) we incorporate absolute-time tokens that specifically handle timestamp knowledge without introducing quantization errors;
(3) we employ a lightweight and high-performance slot-based token compression method to decrease the number of visual tokens to a fixed length, allowing video LLMs to sample more video frames.

In addition to VTG-LLM, we also discovered that existing datasets either suffer from low quality~\citep{zellersluhessel2021merlot} or exhibit significant task imbalance~\citep{ren2023timechat, wang2024hawkeye}. 
To address this issue, we introduce VTG-IT-120K, a dataset comprising 47.2K videos and 120K annotations. We collected this dataset from publicly available sources and re-annotated the low-quality captions using Gemini-1.5 Pro. VTG-IT-120K covers most mainstream VTG tasks, including moment retrieval (63.2K), dense video captioning (37.2K), video summarization (15.2K), and video highlight detection (3.9K). This dataset exhibits a more balanced task distribution compared to existing datasets~\citep{ren2023timechat, wang2024hawkeye}.

Numerical results illustrate the superior performance of the VTG-LLM compared to state-of-the-art video LLM methods. Additionally, ablation studies reveal that absolute-time tokens and sequence-time embedding significantly enhance the model's performance in accurately locating timestamps. Furthermore, the slot-based token compression method outperforms the naive token compression baselines such as sampling and cross-attention in our studied cases.
Our key contributions are summarized as follows:

\begin{itemize}
    \item We present VTG-LLM, a versatile model aims to capture all four VTG tasks. By effectively incorporating timestamp information into video LLMs, VTG-LLM equips these models with the capability to comprehend and process timestamps.Moreover, we have proposed carefully crafted initialization strategies for VTG-LLM to take advantage of pretrained video LLM weights.
    \item We present VTG-IT-120K, a high-quality instruction tuning dataset consisting of 51.9K re-annotated data annotations with superior quality.
    \item Quantitative results demonstrate the superior performance of VTG-LLM across various datasets, such as Charades-STA, QVHighlights, and YouCook2. Our code and datasets are provided in the Supplementary Material.
\end{itemize}

\section{Related Works}

Video Temporal Grounding (VTG) aims to accurately locate the timestamps of events within a given video~\citep{lin2023univtg}. This encompasses tasks such as moment retrieval~\citep{gao2017tall,Zala2023HiREST,oncescu2021queryd,wang2024efficient}, dense video captioning~\citep{zellersluhessel2021merlot,Zala2023HiREST,tang2019coin,caba2015activitynet}, video summarization~\citep{song2015tvsum,gygli2014creating}, and video highlight detection~\citep{lei2021detecting,xiao2023bridging}. Traditional methods mainly address VTG tasks through large-scale video-text pre-training, using training objectives such as video-text contrastive learning~\citep{xu2021videoclip,wang2022internvideo}, video-text matching~\citep{li2023unmasked,chen2024vast}, and masked auto-encoding~\citep{tong2022videomae,zhao2024videoprism}. Although these methods have shown satisfactory results, they require resource-consuming pre-training, lack zero-shot capabilities, and often need further fine-tuning on many downstream tasks.

Large language models (LLMs)~\citep{achiam2023gpt,touvron2023llama} have exhibited considerable potential in capturing knowledge and tackling real-world challenges using a zero-shot approach. Recently, research has explored integrating knowledge from other modalities, such as vision~\citep{liu2024visual} and audio~\citep{ghosal2023text}, to enhance the capabilities of LLMs. Within the visual modality, video large language models (video LLMs) have emerged as a significant research area~\citep{lin2023video,zhang2023video,li2023videochat,song2024moviechat,song2024moviechat+}.
Traditional video LLMs primarily generate captions that summarize videos~\citep{zhang2023video, lin2023video, li2023videochat, maaz2023video, zhu2023minigpt}, but they struggle to accurately pinpoint event timestamps within videos. Some studies have attempted to address this limitation, such as TimeChat~\citep{ren2023timechat}, which constructs time-sensitive instruction tuning datasets and encodes timestamp knowledge into visual tokens. VTimeLLM~\citep{huang2023vtimellm} proposes a LLaVA-like three-stage training method, while LITA~\citep{huang2024lita} introduces fast-slow visual tokens and adds time tokens to LLM tokenizers. Momentor~\citep{qian2024momentor} proposes a time encoder to solve time token quantization errors, and HawkEye~\citep{wang2024hawkeye} constructs a high-quality instruction tuning dataset based on InternVid~\citep{wang2022internvideo} for the moment retrieval task.
NumPro~\citep{wu2024number} add frame numbers to frames for ease of temporal understanding.
However, existing instruction tuning datasets for VTG tasks are often low-quality or exhibit extreme task imbalance, which hinders model performance across tasks. To address this issue, we propose a high-quality dataset, VTG-IT, and introduce VTG-LLM, a model comprising three well-designed components to enhance video LLM performance on VTG tasks.
\section{Method}

In this section, we introduce the VTG-LLM model and the VTG-IT-120K dataset. The overview of the VTG-LLM model can be found in Figure~\ref{fig:overview}. 

\subsection{Overview of VTG-LLM}

In this subsection, we present the detailed structure of VTG-LLM, which comprises three key components designed to enhance video LLMs' understanding of timestamps. Specifically, we propose: (1) a sequence-time embedding mechanism that directly incorporates timestamp information into visual tokens; (2) the introduction of absolute-time tokens without quantization errors to differentiate time-related knowledge from other digit-related knowledge; and (3) the implementation of slot-based token compression to enable video LLMs to effectively process more frames. We will discuss these components in detail throughout this section.

\subsection{Sequence-Time Embedding}

\paragraph{Sequence embedding.} Existing studies~\citep{zhang2023video,ren2023timechat} employ sequence embedding to incorporate relative time information into visual tokens. Specially, given $N \times M$
tokens $\{\zz_{i,j} | 1 \le i \le N, 1 \le j \le M \}$ sampled from $N$ frames, the process of adding sequence embedding onto visual tokens can then be represented by:
\begin{align}
    \hat{\zz}_{i,j} = \zz_{i,j} + [\mW_{s}]_{i} \, , 
\end{align}
where $\mW_{s} \in \R^{N \times d}$ is the weight of sequence-embedding matrix.

\paragraph{Sequence-Time Embedding.} Although promising, the sequence embedding only contains information about the temporal order and may not accurately represent knowledge about the absolute timestamps. For instance, sampled frames may not be uniformly distributed throughout the entire video~\citep{ren2023timechat,huang2023vtimellm}. Moreover, the sampling intervals can vary significantly for videos of different lengths. These issues make it difficult to infer the timestamps of the frames simply using sequence embedding. 
To address this issue, in addition to the sequence embedding, we also add the absolute time embedding as shown in the following equation
\begin{align}
    \hat{\zz}_{i,j} = \zz_{i,j} + [\mW_{s}]_{i} + [\mW_{t}]_{t} \, .
\end{align}
Here, $t$ is the absolute timestamp (in seconds) of the corresponding frames. $\mW_{t} \in \R^{T \times d}$ is the absolute time embedding. $T$ is the maximum timestamp (in seconds). \textit{It's important to note that to avoid disrupting the visual tokens generated by pre-trained visual encoders and to speed up the convergence, $\mW_{t}$ is initialized by setting all entries to zero.} 
Since we employ pretrained vision encoder and Qformer modules, the performance of VTG-LLM is significantly degraded without the zero-initialization method. This is evident in the "TE Random Initialize" section of Table~\ref{tab:ablation}.

Due to the imbalanced nature of video lengths and the gap between training and test data, there may be timestamps for which the absolute time embedding has not been trained. To address this issue, we further introduce a test-time interpolation mechanism. In detail, for the timestamp $t$ which is not trained during training, we first find the timestamps $t_l$ and $t_r$ that satisfy:
\begin{align}
    t_l = \argmax_{t_l} t_1 < t, t_l \in \cT_{tr} \, ,  t_r = \argmin_{t_r} t_r > t, t_r \in \cT_{tr} \, ,
\end{align}
where $\cT_{tr}$ is the set of timestamps with trained absolute-time embedding. Then the absolute-time embedding of timestamp $t$ is given by 
\begin{align}
    [\mW_{t}]_t = \frac{t - t_l}{(t_r - t_l)} [\mW_{t}]_{t_l} + \frac{t_r - t}{(t_r - t_l)} [\mW_{t}]_{t_r} \, .
\end{align}
Please refer to Appendix~\citep{guo2024vtg} for ablation studies on using test-time interpolation.

\paragraph{Discussion on existing techniques that integrating information of timestamps into visual tokens.}
In addition to incorporating time embeddings into visual tokens, there are other techniques for integrating timestamp information into visual tokens. For instance, adding text inputs into Qformer~\citep{ren2023timechat} and inserting time tokens before visual tokens~\citep{hua2024v2xum}. However, we believe that these techniques are orthogonal to the time embedding approach, and it is also possible to combine these approaches in future work.

\subsection{Absolute-Time Tokens}

The incorporation of unique time tokens in the tokenizer has shown benefits in video temporal grounding tasks, as evidenced by various studies~\citep{yang2023vid2seq,qian2024momentor,huang2024lita}. However, using relative time tokens (frame ID) exposes some limitations. First, quantization errors grow linearly with video length, complicating fine-grained predictions for longer videos. Second, training token embedding from scratch hinders leveraging pretrained video LLMs' benefits. Our goal is to create a novel time token mechanism that accurately represents fine-grained timestamps and easily adapts to pretrained video LLMs.

\paragraph{Absolute-time tokens for resolving the quantization errors.} To resolve the issue of quantization errors, we introduce the concept of \textit{absolute-time tokens}. As depicted in Figure~\ref{fig:overview}, we have incorporated eleven time tokens into the tokenizer. These consist of ten digit time tokens representing the digits from 0 to 9, and an additional token for a decimal point. All timestamps are represented using six time tokens. For instance, the time 120.5 seconds would be formatted as $\langle t_0 \rangle\langle t_1 \rangle\langle t_2 \rangle\langle t_0 \rangle\langle t_{dot} \rangle\langle t_5 \rangle$. This formulation allows the time tokens to handle videos more than 1 hours in length, with the precision remaining constant regardless of video length increases. Importantly, it is essential to format all timestamps using the same number of time tokens. As shown in the "Time Token not Formatted" section of Table~\ref{tab:ablation}, we found that maintaining a consistent format for all timestamps significantly improves model performance.

\paragraph{Initialization of token embedding for absolute-time tokens.} While using absolute time tokens eliminates quantization errors, we found that randomly initialized time tokens adversely affect the original token embedding space, hindering LLMs from learning precise time token knowledge. Consequently, this leads to subpar model performance, as demonstrated in the "Token Embedding not Initialized" section of Table~\ref{tab:ablation}.
To tackle this issue, our goal is to initialize the time-related knowledge using the digit-related knowledge, as the digit-related knowledge has already been well-trained during the LLM pretraining stage.

This is achieved by adjusting the initialization of the weights of token embedding and LLM prediction head. For example, consider the token embedding matrix $\mW_{token}$, when initializing the embedding for the time token "$\langle t_1 \rangle$", we align the embedding of this time token with the embedding of the token "1":
\begin{align}
    [\mW_{token}]_{\text{ID}(\text{'$\langle t_1 \rangle$'}), j} = [\mW_{token}]_{\text{ID}(\text{'1'}), j} \, , \forall 0 \le j < d \, ,
\end{align}
where ID() denotes the token ID of the given token string. Similarly, for the LM prediction head, we employ the same method to transfer the knowledge from the digit tokens to the time tokens.

\setlength{\parskip}{0.5pt plus0.5pt minus0.5pt}

\subsection{Slot-Based Token Compression} 

Compressing visual tokens is crucial in VTG tasks.
On one hand, models cannot produce reliable predictions without adequate visual input. On the other hand, the context length of LLMs inherently imposes a limitation.
To address this, we propose a straightforward yet efficient approach called slot-based token compression, which compresses visual tokens to a fixed number, thereby enabling models to sample more frames.


\begin{figure}
    \centering
    \includegraphics[width=1.\linewidth]{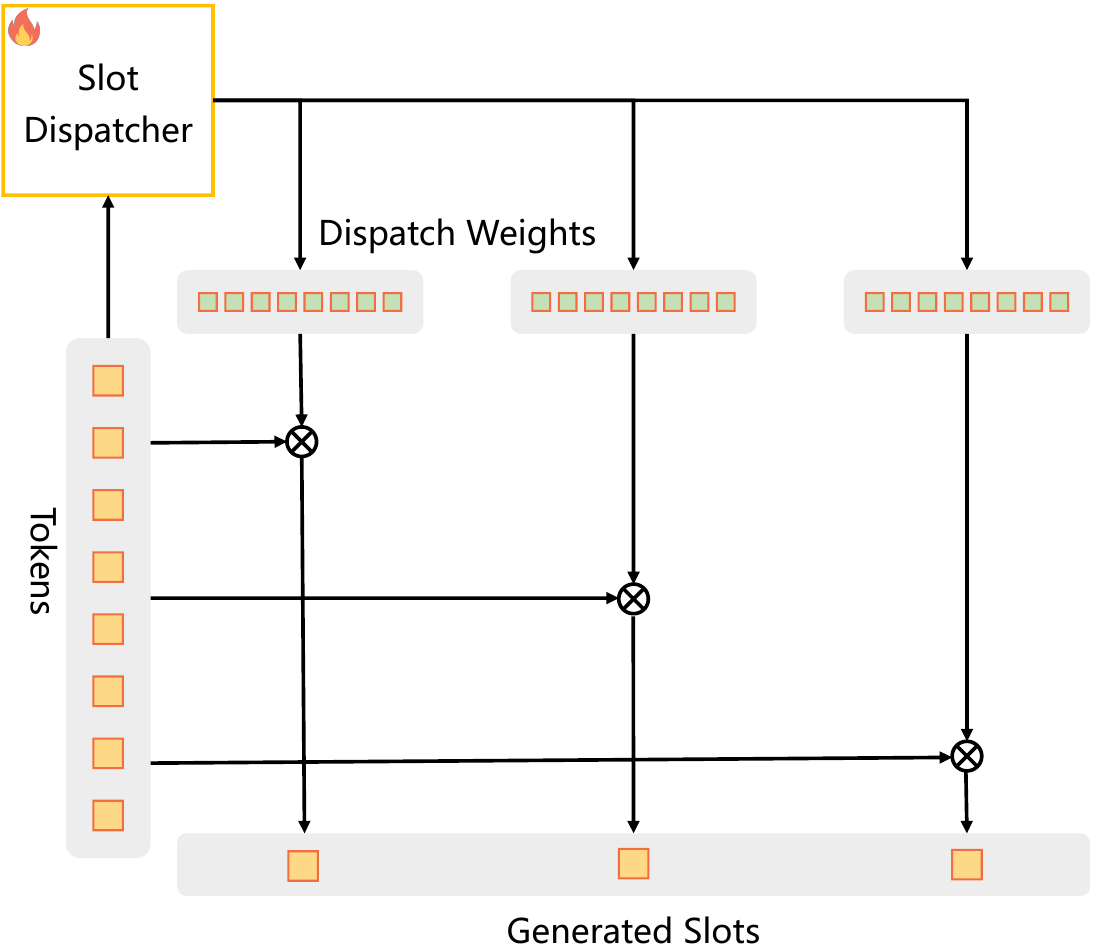}
    \caption{Overview of slot based token compression.}
    \label{fig:slot-sampler}
\end{figure}


    

As illustrated in Figure~\ref{fig:slot-sampler}, given $N$ visual tokens $\zz_1, \cdots, \zz_N$, and the trainable slot dispatcher $\mathbf{\Phi} \in \R^{K \times d}$, where $K$ represents the number of slots, the visual tokens first pass through the slot dispatcher to obtain the dispatch weights, and the tokens are mixed based on the dispatch weights to generate the slots. For instance, the slot $k$ is generated by
\begin{align}
    \ss_k = \sum_{i=1}^{N} \frac{\exp(\mathbf{\Phi}_{k}^{T} \zz_i)}{\sum_{j=1}^{N} \exp(\mathbf{\Phi}_{k}^{T} \zz_j)} \zz_i \, .
\end{align}
The slots are subsequently fed into the visual projection layers and serve as visual tokens.

\paragraph{Discussion on slot mechanism}
    The slot-based token compression is inspired by the slot mechanism in SoftMoE~\citep{puigcerver2023sparse}. However, in SoftMoE, the slots are not used to reduce sequence length, which is distinct from our primary objective. Moreover, some existing multi-modal LLMs compress the token number through cross-attention~\citep{zhang2023video,bai2023qwen,ren2023timechat}. Our approach involves training only one matrix $\mathbf{\Phi}$, making it more computationally efficient and less data-consuming, and thereby achieve better performance (Table~\ref{tab:ablation}) on relatively small-scale instruction tuning datasets.

\begin{table*}[!t]
    \centering
        \begin{tabular}{l c c c c c c c c c}
            \toprule
            \multirow{2}{*}{Model} & \multicolumn{3}{c}{Youcook2}                         & \multicolumn{2}{c}{Charades-STA}                      & \multicolumn{2}{c}{QVHighlights}                                                                                                                                                                                                                                    \\
            \cmidrule(lr){2-4} \cmidrule(lr){5-6} \cmidrule(lr){7-8}
                                        & SODA\_c                                                          & CIDEr                                                         & F1 Score                                                          & $\text{R@1}_{\text{(IOU=0.5)}}$                                                         & $\text{R@1}_{\text{(IOU=0.7)}}$                                                                   & mAP & HIT@1                                                                 \\
            \midrule
            \textit{\textbf{\small Traditional Video LLMs}} \\
            Valley (7B) & 0.1 & 0.0 & 1.5 & 4.7 & 1.6 & 10.9 & 15.2                   \\
            VideoChat (7B) & 0.2 & 0.6 & 3.4 & 3.2 & 1.4 & 13.1 & 18.1 \\
            Video-LLaMA (7B) & 0.0 & 0.0 & 0.1 & 2.7 & 1.2 & 11.3 & 15.6 \\
            Video-ChatGPT (7B) &  &  &  & 7.7 & 1.7 & 3.8 &  \\
            \midrule
            \textit{\textbf{\small Temporal Grounding Video LLMs}} \\
            TimeChat (7B) & 1.2 & 3.4 & 12.6 & 32.2 & 13.4 & 14.5 & 23.9\\
            VTimeLLM (7B) & 1.0  & 3.6  & 9.1  & 27.5 & 11.4 &  & \\
            Momentor (7B) &  &  &  & 26.6 & 11.6 & 7.6 &  \\
            HawkEye (7B) &  &  &  & 31.4 & 14.5 &  & \\
            \midrule
            VTG-LLM (7B) & \textbf{1.5} & \textbf{5.0} & \textbf{17.5} & \textbf{33.8} & \textbf{15.7} & \textbf{16.5} & \textbf{33.5} \\
            \bottomrule
        \end{tabular}%
    \caption{
        Zero-shot performance of algorithms over various tasks.
    }
    \label{tab:zero-shot-vtg}
\end{table*}

\subsection{VTG-IT-120K: Formatted Time-Sensitive Instruction Tuning Dataset}
\begin{figure}
    \centering
    \includegraphics[width=1.\linewidth]{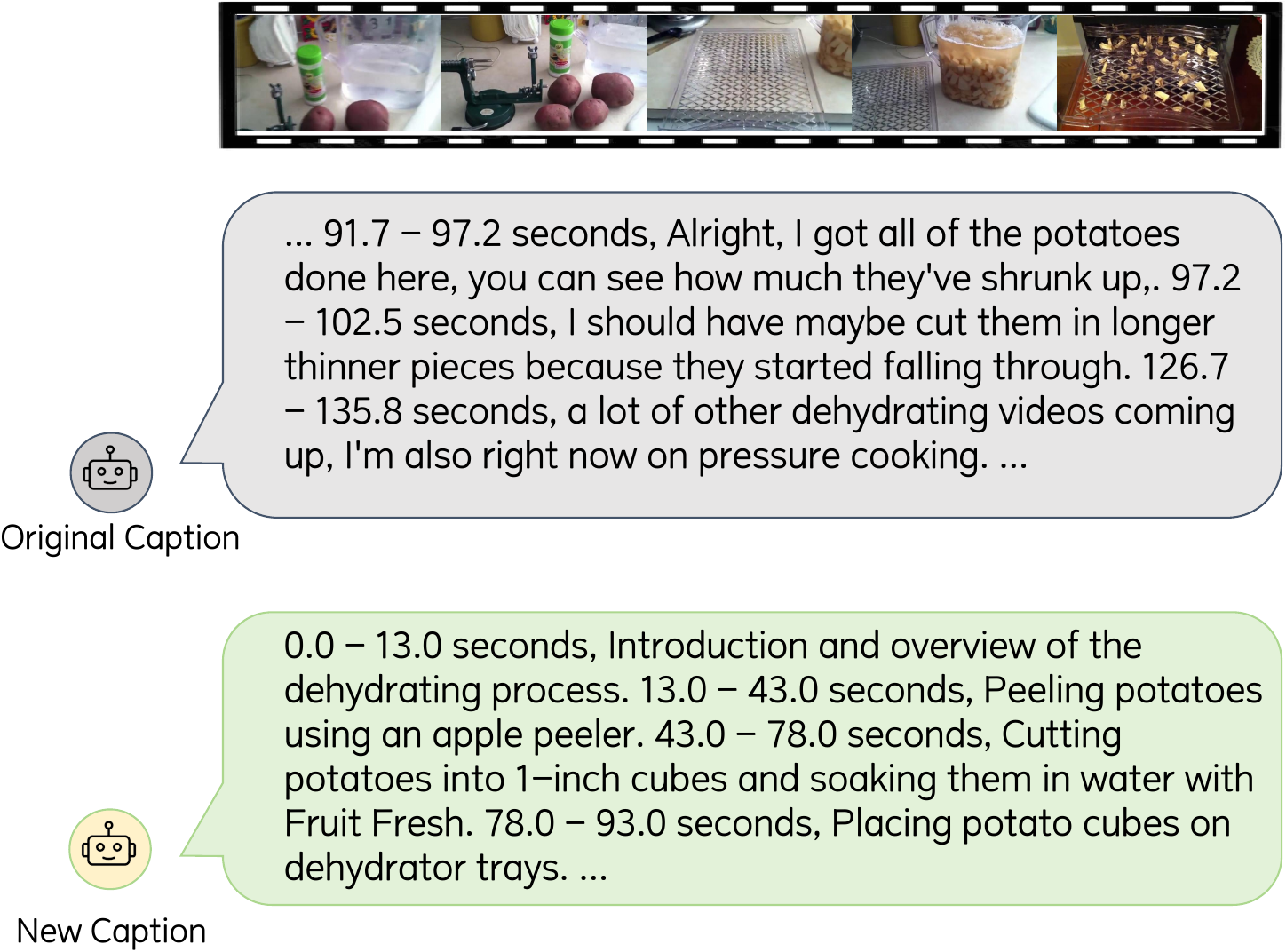}
    \caption{Example of new annotations.}
    \label{fig:annotation-example}
\end{figure}

In this section, we introduce VTG-IT-120K, a dataset comprising 120K publicly available video-text pairs, building on the TimeIT dataset~\citep{ren2023timechat}. Additionally, we have re-annotated 51.9K low-quality video annotations using Gemini 1.5-Pro~\footnote{The details of the annotation process can be found in Appendix~\citep{guo2024vtg}.}.
A comparison between the original and new annotations can be seen in Figure~\ref{fig:annotation-example}. It is clear that the revised captions are more succinct and contain less information unrelated to the visual content.
The VTG-IT-120K dataset encompasses four distinct video temporal grounding tasks.
\begin{itemize}
    \item \textit{Moment Retrieval (63.2K):} For the moment retrieval task, we use HiREST$_{grounding}$~\citep{zala2023hierarchical}, QuerYD~\citep{oncescu2021queryd}, DiDeMo~\citep{hendricks2018localizing}, and VTG-IT-MR.
    \item \textit{Dense Video Captioning (37.2K):} For the dense video captioning task, we use HiREST$_{step}$~\citep{zala2023hierarchical}, COIN~\citep{tang2019coin}, ActivityNet Captions~\citep{caba2015activitynet}, and VTG-IT-DVC.
    \item \textit{Video Summarization (15.2K):} For the video summarization task, we use TVSum~\citep{song2015tvsum}, SumMe~\citep{gygli2014creating}, and VTG-IT-VS.
    \item \textit{Video Highlight Detection (3.9K):} For the video highlight detection task, we use VTG-IT-VHD.
\end{itemize}
The VTG-IT-X annotations for the four tasks are re-annotated using 16K videos from the YT-Temporal-180M dataset~\citep{zellersluhessel2021merlot}.

\paragraph{Data formatting.} All tasks are structured as QA pairs, with varied questions designed to help models comprehend human intent. The answers are formatted to facilitate knowledge transfer between different tasks. Detailed examples are provided in the Appendix~\citep{guo2024vtg}.

\section{Experiments}

In the experimental section, we primarily aim to address the following questions~\footnote{Fine-tuned performance, results on ActivityNet Captions, more ablation studies, qualitative comparisons, and case studies can be found in Appendix~\citep{guo2024vtg}.}:

\begin{itemize}
    \item Q1. Will VTG-LLM achieve better zero-shot performance compared to SOTA video LLMs?
    \item Q2. Is it necessary to use time embeddings, and what will happen if time embeddings are randomly initialized?
    \item Q3. How will the special time tokens affect the model's performance?
    \item Q4. Will the slot-based compression method outperform other compression methods?
\end{itemize}

\subsection{Experiment Settings} 

\label{sec:experiment-setting}

\paragraph{Models and training configuration.} Unless otherwise specified, we employ ViT-G/14 from EVA-CLIP~\citep{sun2023eva} and the Qformer from InstructBLIP~\citep{dai2024instructblip} as the visual backbone. The vision projection layer and the weights of sequence embedding are initialized using the Video-LLaMA checkpoint~\citep{zhang2023video}. For the language model backbone, we utilize LLaMA-2-7B~\citep{touvron2023llama}. The number of slots is set to 256.  The models are trained on the VTG-IT-120k and a randomly sampled subset (97k) from the Valley dataset~\citep{luo2023valley}. 
For the training process, we initially divide the videos uniformly into 96 segments, from each of which we then randomly select one frame. For the testing process, we uniformly select 96 frames from the entire video. Training is carried out using 16 ATN 910B NPUs with a batch size of 64. The fine-tuning also uses 16 NPUs and set the batch size to 16.
We set the learning rate to 3e-5 initially, and train the models for 10 epochs by default.

\paragraph{Evaluation datasets, metrics, and baseline models.} We evaluate the model performance on three different tasks:
\begin{itemize}
    \item \textit{dense video captioning.} 
    We employ Youcook2~\citep{zhou2018towards} as the test dataset, and following the evaluation settings in TimeChat~\citep{ren2023timechat}. Metrics including CIDEr~\citep{vedantam2015cider}, which assessing the quality of the captions; (2) SODA\_c~\citep{fujita2020soda} for story-level evaluation; (3) F1 score to measure the model's ability to accurately locate timestamps;
    \item \textit{Moment retrieval.}
    We utilize test set of Charades-STA~\citep{gao2017tall} for the moment retrieval task and report the recall at IOU thresholds of 0.5 and 0.7.
    \item \textit{Video highlight detection.} We employ the validation set of the QVHighlights dataset~\citep{lei2021detecting} and report the mean average precision (mAP) with IOU thresholds of 0.5 and 0.75, as well as the HIT@1, which represents the hit ratio of the highest scored clip.\looseness=-1
\end{itemize}
For baseline models, we select Valley~\citep{luo2023valley}, VideoChat~\citep{li2023videochat}, Video-ChatGPT~\citep{maaz2023video}, and Video-LLaMA~\citep{zhang2023video} as examples of traditional video LLMs. For video LLMs specifically designed for VTG tasks, we choose TimeChat~\citep{ren2023timechat}, VTimeLLM~\citep{huang2023vtimellm}, Momentor~\citep{qian2024momentor}, and HawkEye~\citep{wang2024hawkeye}~\footnote{Although related, LITA~\citep{huang2024lita} employs self-built evaluation sets, making fair comparison difficult.}.\looseness=-1

\subsection{Numerical Results}

\paragraph{R1. VTG-LLM achieves superior zero-shot performance on various VTG tasks.} In Table~\ref{tab:zero-shot-vtg}, we evaluate the performance of the VTG-LLM on three VTG tasks, including dense video captioning (Youcook2), moment retrieval (Charades-STA), and video highlight detection (QVHighlights). The results indicate that: (1) The VTG-LLM model outperforms all existing 7B video LLM models on the zero-shot setting across all three datasets; 
(2) The improvement of the VTG-LLM on metrics that evaluate timestamp location accuracy, such as the F1 score for the Youcook2 dataset and Recall for the Charades-STA dataset, is more pronounced. This suggests that the performance gain of the VTG-LLM primarily stems from more accurate timestamp location. Additionally, performance on other metrics like CIDEr and SODA\_c also improves.
(3) As shown in Table~\ref{tab:performance-wrt-duration}, the performance of VTG-LLM remains relatively robust across different video durations.

\begin{table*}[!t]
    \centering
        \begin{tabular}{l c c c c c c c c c}
            \toprule
            \multirow{2}{*}{Model} & \multicolumn{3}{c}{Youcook2}                         & \multicolumn{2}{c}{Charades-STA}                      & \multicolumn{2}{c}{QVHighlights}                                                                                                                                                                                                                                    \\
            \cmidrule(lr){2-4} \cmidrule(lr){5-6} \cmidrule(lr){7-8}
                                        & SODA\_c                                                          & CIDEr                                                         & F1 Score                                                          & $\text{R@1}_{\text{(IOU=0.5)}}$                                                         & $\text{R@1}_{\text{(IOU=0.7)}}$                                                                   & mAP & HIT@1                                                                 \\
            \midrule
            \textit{Ablation Studies on data} \\
            VTG-LLM (TimeIT) & 1.0 & 2.8 & 12.0 & 35.1 & 14.8 & 14.9 & 19.1 \\
            \midrule
            \textit{Ablation Studies on STE} \\
            SE Only & 1.4 & 4.4 & 17.2 & \underline{33.6} & \underline{14.8} & \underline{16.4} & 32.8 \\
            SE + TE (Random Initialize) & 1.3 & 4.2 & 16.8 & 21.4 & 10.5 & 14.2 & 22.5 \\
            \midrule
            \textit{Ablation Studies on Time Token} \\
            No Time Token & \underline{1.7} & \underline{5.5} & 17.4 & 29.5 & 13.6 & 16.2 & \underline{33.9} \\
            Time Token not Formatted & 1.5 & 5.1 & 16.8 & 27.0 & 12.4 & 15.0 & 28.8\\
            Token Embedding not Initialized & 1.4 & 4.8 & 16.1 & 15.6 & 6.5 & 14.0 & 21.9\\
            \midrule
            \textit{Ablation Studies on Token Compression} \\
            Entropy Sampling & 1.3 & 4.2 & 16.8 & 21.1 & 9.6 & 14.0 & 20.1\\
            Diverse Sampling & 1.3 & 4.4 & 16.8 & 22.4 & 10.5 & 14.2 & 22.7\\
            Cross Attention & 1.3 & 4.2 & \underline{17.6} & 19.9 & 8.8 & 13.7 & 21.0\\
            \midrule
            Original VTG-LLM & 1.5 & 5.0 & 17.5 & 33.8 & 15.7 & 16.5 & 33.5 \\
            \bottomrule
        \end{tabular}%
    \caption{
        Ablation studies on VTG-LLM.
    }
    \label{tab:ablation}
\end{table*}

\begin{table}[!t]
    \centering
    \begin{tabular}{l c c c c c c c c c}
    \toprule
    YouCook2 & CIDEr & METEOR & F1 & SODA\_c \\
    \midrule
    $[0s, 180s)$ & 6.3 & 2.3 & 20.0 & 1.9 \\
    $[180s, 240s)$ & 4.6 & 1.8 & 18.8 & 1.4 \\
    $[240s, 300s)$ & 5.0 & 1.9 & 16.6 & 1.6 \\
    $[300s, 420s)$ & 5.6 & 2.1 & 17.9 & 1.5 \\
    $[420s, inf)$ & 4.1 & 1.6 & 15.1 & 1.3 \\
    \bottomrule
    \end{tabular}
    \caption{
        Performance with different video durations.
    }
    \label{tab:performance-wrt-duration}
\end{table}

\paragraph{R2.Time embedding boost the model performance when using zero initialization technique.} 
In the "Ablation Studies on STE" section of Table~\ref{tab:ablation}, we show the performance of VTG-LLM when only using sequence embedding ("SE Only") and using sequence-time embedding but randomly initializing the weights of time-embedding ("TE Random Initialize"). The results indicate that: (1) the time-embedding is necessary for VTG-LLM to achieve the best result. Without the time-embedding, the performance of VTG-LLM on all datasets decreases; (2) Randomly initializing the time embedding significantly hurts the performance of VTG-LLM.

\paragraph{R3. Special time tokens significantly boost temporal grounding capacity with the cost of slightly caption quality reduction.} In the "Ablation Studies on Time Token" section of Table~\ref{tab:ablation}, we show the performance of VTG-LLM when not using special time tokens ("No Time Token"), when not formatting the length of timestamps to six time tokens ("Time Token not Formatted"), and when do not initialize the token embedding and LLM head for time tokens ("Token Embedding not Initialized"). The results indicate that
\begin{itemize}
    \item Using time tokens significantly boosts the model's capacity to accurately locate the timestamps, improving the score on metrics like F1 score, $\text{R@1}_{\text{(IOU=0.5)}}$, $\text{R@1}_{\text{(IOU=0.7)}}$, and mAP. However, the caption-quality related metrics like SODA\_c and CIDEr slightly decrease. This suggests that there exists a trade-off between caption quality and timestamp accuracy when using time token.\looseness=-1
    \item We found that the "No Time Token" setting tends to predict fewer timestamps on QVHighlights dataset, which may lead to better HIT@1 performance but a worse mAP score.\looseness=-1
    \item Formatting the timestamps using the same number of time tokens significantly enhances the model performance ("Time Token not Formatted"). We believe that this is because the formatted setting simplifies the task, making the models needless to identify the number of tokens in timestamps.\looseness=-1
    \item Initialize the weights of token embedding and LLM prediction head is essential for time token to work well.\looseness=-1 
\end{itemize}
\textit{In summary, we recommend using special time tokens to achieve significant improvements in the accuracy of locating timestamps, which is the primary goal of this paper.} Moreover, investigating the trade-off between caption quality and timestamp location accuracy in future work would be interesting.\looseness=-1

\paragraph{R4. Slot-based token compression outperform naive compression baselines.} 
In the "Ablation Studies on Token Compression" section of Table~\ref{tab:ablation}, we present the performance of VTG-LLM using different token compression methods and include some naive baselines. The "Entropy Sampling" method involves sampling the number of tokens to 256 by selecting tokens with the maximum entropy, where the entropy is estimated using k-nearest neighbor distances~\citep{kozachenko1987sample}. The "Diverse Sampling" method involves sampling tokens to 256 using k-means++~\citep{arthur2007k} to select tokens with the maximum distance. The "Cross Attention" method involves using a cross-attention layer to compress the number of tokens. The results indicate that: (1) The slot-based sampling achieves better performance among all compression methods; (2) Sampling-based methods perform poor compared to slot-based token compression method. We conjecture that this is because the sampling methods drop too much necessary information; (3) The cross-attention method performs even worse than sampling-based methods, which might be attributed to the attention mechanism requiring a large amount of data to perform well.\looseness=-1

\section{Conclusion, Limitation, and Future Works}

\label{sec:conclusion}

In this paper, we propose the VTG-LLM, an improved video LLM model for integrating knowledge about timestamps, enhancing the zero shot performance on four downstream VTG tasks.
Extensive numerical results demonstrate the superior zero-shot performance of VTG-LLM over state-of-the-art video LLM models on Charades-STA, QVHighlights, and Youcook2 datasets, with each proposed component contributing to individual performance gains. 

Nonetheless, there are certain limitations in the VTG-LLM that necessitate additional research. For example, while this study mainly concentrates on the accuracy of timestamp locations, it would be advantageous to explore methods for improving other aspects, such as the precision of salient scores and the quality of captions.
Furthermore, we did not incorporate the audio components of the videos in this study. Moreover, evaluating the performance of slot-based compression on a more diverse set of tasks is also worth investigating.
\looseness=-1

\section*{Acknowledgments}
This work is supported in part by the funding from Shenzhen Institute of Artificial Intelligence and Robotics for Society, in part by the Shenzhen Key Lab of Crowd Intelligence Empowered Low-Carbon Energy Network (Grant No. ZDSYS20220606100601002) , in part by Shenzhen Stability Science Program 2023, and in part by the Guangdong Provincial Key Laboratory of Future Networks of Intelligence (Grant No. 2022B1212010001).

\bibliography{aaai25}

\begin{thebibliography}{55}
\providecommand{\natexlab}[1]{#1}

\bibitem[{Achiam et~al.(2023)Achiam, Adler, Agarwal, Ahmad, Akkaya, Aleman,
  Almeida, Altenschmidt, Altman, Anadkat et~al.}]{achiam2023gpt}
Achiam, J.; Adler, S.; Agarwal, S.; Ahmad, L.; Akkaya, I.; Aleman, F.~L.;
  Almeida, D.; Altenschmidt, J.; Altman, S.; Anadkat, S.; et~al. 2023.
\newblock Gpt-4 technical report.
\newblock \emph{arXiv preprint arXiv:2303.08774}.

\bibitem[{Arthur, Vassilvitskii et~al.(2007)}]{arthur2007k}
Arthur, D.; Vassilvitskii, S.; et~al. 2007.
\newblock k-means++: The advantages of careful seeding.
\newblock In \emph{Soda}, volume~7, 1027--1035.

\bibitem[{Bai et~al.(2023)Bai, Bai, Yang, Wang, Tan, Wang, Lin, Zhou, and
  Zhou}]{bai2023qwen}
Bai, J.; Bai, S.; Yang, S.; Wang, S.; Tan, S.; Wang, P.; Lin, J.; Zhou, C.; and
  Zhou, J. 2023.
\newblock Qwen-vl: A frontier large vision-language model with versatile
  abilities.
\newblock \emph{arXiv preprint arXiv:2308.12966}.

\bibitem[{Chen et~al.(2024)Chen, Li, Wang, Zhao, Sun, Zhu, and
  Liu}]{chen2024vast}
Chen, S.; Li, H.; Wang, Q.; Zhao, Z.; Sun, M.; Zhu, X.; and Liu, J. 2024.
\newblock Vast: A vision-audio-subtitle-text omni-modality foundation model and
  dataset.
\newblock \emph{Advances in Neural Information Processing Systems}, 36.

\bibitem[{Cheng et~al.(2024)Cheng, Leng, Zhang, Xin, Li, Chen, Zhu, Zhang, Luo,
  Zhao et~al.}]{cheng2024videollama}
Cheng, Z.; Leng, S.; Zhang, H.; Xin, Y.; Li, X.; Chen, G.; Zhu, Y.; Zhang, W.;
  Luo, Z.; Zhao, D.; et~al. 2024.
\newblock VideoLLaMA 2: Advancing Spatial-Temporal Modeling and Audio
  Understanding in Video-LLMs.
\newblock \emph{arXiv preprint arXiv:2406.07476}.

\bibitem[{Dai et~al.(2024)Dai, Li, Li, Tiong, Zhao, Wang, Li, Fung, and
  Hoi}]{dai2024instructblip}
Dai, W.; Li, J.; Li, D.; Tiong, A. M.~H.; Zhao, J.; Wang, W.; Li, B.; Fung,
  P.~N.; and Hoi, S. 2024.
\newblock Instructblip: Towards general-purpose vision-language models with
  instruction tuning.
\newblock \emph{Advances in Neural Information Processing Systems}, 36.

\bibitem[{Fabian Caba~Heilbron and Niebles(2015)}]{caba2015activitynet}
Fabian Caba~Heilbron, B.~G., Victor~Escorcia; and Niebles, J.~C. 2015.
\newblock ActivityNet: A Large-Scale Video Benchmark for Human Activity
  Understanding.
\newblock In \emph{Proceedings of the IEEE Conference on Computer Vision and
  Pattern Recognition}, 961--970.

\bibitem[{Fujita et~al.(2020)Fujita, Hirao, Kamigaito, Okumura, and
  Nagata}]{fujita2020soda}
Fujita, S.; Hirao, T.; Kamigaito, H.; Okumura, M.; and Nagata, M. 2020.
\newblock SODA: Story oriented dense video captioning evaluation framework.
\newblock In \emph{Computer Vision--ECCV 2020: 16th European Conference,
  Glasgow, UK, August 23--28, 2020, Proceedings, Part VI 16}, 517--531.
  Springer.

\bibitem[{Gao et~al.(2017)Gao, Sun, Yang, and Nevatia}]{gao2017tall}
Gao, J.; Sun, C.; Yang, Z.; and Nevatia, R. 2017.
\newblock Tall: Temporal activity localization via language query.
\newblock In \emph{Proceedings of the IEEE international conference on computer
  vision}, 5267--5275.

\bibitem[{Ghosal et~al.(2023)Ghosal, Majumder, Mehrish, and
  Poria}]{ghosal2023text}
Ghosal, D.; Majumder, N.; Mehrish, A.; and Poria, S. 2023.
\newblock Text-to-audio generation using instruction-tuned llm and latent
  diffusion model.
\newblock \emph{arXiv preprint arXiv:2304.13731}.

\bibitem[{Guo et~al.(2024)Guo, Liu, Li, Tang, Chen, and Zhao}]{guo2024vtg}
Guo, Y.; Liu, J.; Li, M.; Tang, X.; Chen, X.; and Zhao, B. 2024.
\newblock VTG-LLM: Integrating Timestamp Knowledge into Video LLMs for Enhanced
  Video Temporal Grounding.
\newblock \emph{arXiv preprint arXiv:2405.13382}.

\bibitem[{Gygli et~al.(2014)Gygli, Grabner, Riemenschneider, and
  Van~Gool}]{gygli2014creating}
Gygli, M.; Grabner, H.; Riemenschneider, H.; and Van~Gool, L. 2014.
\newblock Creating summaries from user videos.
\newblock In \emph{Computer Vision--ECCV 2014: 13th European Conference,
  Zurich, Switzerland, September 6-12, 2014, Proceedings, Part VII 13},
  505--520. Springer.

\bibitem[{Han et~al.(2024)Han, Seo, Park, Nam, and Kwak}]{han2024unleash}
Han, D.; Seo, S.; Park, E.; Nam, S.-U.; and Kwak, N. 2024.
\newblock Unleash the Potential of CLIP for Video Highlight Detection.
\newblock \emph{arXiv preprint arXiv:2404.01745}.

\bibitem[{Hendricks et~al.(2018)Hendricks, Wang, Shechtman, Sivic, Darrell, and
  Russell}]{hendricks2018localizing}
Hendricks, L.~A.; Wang, O.; Shechtman, E.; Sivic, J.; Darrell, T.; and Russell,
  B. 2018.
\newblock Localizing Moments in Video with Temporal Language.
\newblock In \emph{Empirical Methods in Natural Language Processing (EMNLP)}.

\bibitem[{Hua et~al.(2024)Hua, Tang, Xu, and Luo}]{hua2024v2xum}
Hua, H.; Tang, Y.; Xu, C.; and Luo, J. 2024.
\newblock V2Xum-LLM: Cross-Modal Video Summarization with Temporal Prompt
  Instruction Tuning.
\newblock \emph{arXiv preprint arXiv:2404.12353}.

\bibitem[{Huang et~al.(2023)Huang, Wang, Chen, Song, and
  Zhu}]{huang2023vtimellm}
Huang, B.; Wang, X.; Chen, H.; Song, Z.; and Zhu, W. 2023.
\newblock Vtimellm: Empower llm to grasp video moments.
\newblock \emph{arXiv preprint arXiv:2311.18445}, 2(3): 9.

\bibitem[{Huang et~al.(2024)Huang, Liao, Radhakrishnan, Yin, Molchanov, Yu, and
  Kautz}]{huang2024lita}
Huang, D.-A.; Liao, S.; Radhakrishnan, S.; Yin, H.; Molchanov, P.; Yu, Z.; and
  Kautz, J. 2024.
\newblock LITA: Language Instructed Temporal-Localization Assistant.
\newblock \emph{arXiv preprint arXiv:2403.19046}.

\bibitem[{Kozachenko and Leonenko(1987)}]{kozachenko1987sample}
Kozachenko, L.~F.; and Leonenko, N.~N. 1987.
\newblock Sample estimate of the entropy of a random vector.
\newblock \emph{Problemy Peredachi Informatsii}, 23(2): 9--16.

\bibitem[{Lei, Berg, and Bansal(2021)}]{lei2021detecting}
Lei, J.; Berg, T.~L.; and Bansal, M. 2021.
\newblock Detecting moments and highlights in videos via natural language
  queries.
\newblock \emph{Advances in Neural Information Processing Systems}, 34:
  11846--11858.

\bibitem[{Li et~al.(2023{\natexlab{a}})Li, He, Wang, Li, Wang, Luo, Wang, Wang,
  and Qiao}]{li2023videochat}
Li, K.; He, Y.; Wang, Y.; Li, Y.; Wang, W.; Luo, P.; Wang, Y.; Wang, L.; and
  Qiao, Y. 2023{\natexlab{a}}.
\newblock VideoChat: Chat-Centric Video Understanding.
\newblock \emph{arXiv preprint arXiv:2305.06355}.

\bibitem[{Li et~al.(2023{\natexlab{b}})Li, Wang, Li, Wang, He, Wang, and
  Qiao}]{li2023unmasked}
Li, K.; Wang, Y.; Li, Y.; Wang, Y.; He, Y.; Wang, L.; and Qiao, Y.
  2023{\natexlab{b}}.
\newblock Unmasked teacher: Towards training-efficient video foundation models.
\newblock In \emph{Proceedings of the IEEE/CVF International Conference on
  Computer Vision}, 19948--19960.

\bibitem[{Lin et~al.(2023{\natexlab{a}})Lin, Zhu, Ye, Ning, Jin, and
  Yuan}]{lin2023video}
Lin, B.; Zhu, B.; Ye, Y.; Ning, M.; Jin, P.; and Yuan, L. 2023{\natexlab{a}}.
\newblock Video-llava: Learning united visual representation by alignment
  before projection.
\newblock \emph{arXiv preprint arXiv:2311.10122}.

\bibitem[{Lin et~al.(2023{\natexlab{b}})Lin, Zhang, Chen, Pramanick, Gao, Wang,
  Yan, and Shou}]{lin2023univtg}
Lin, K.~Q.; Zhang, P.; Chen, J.; Pramanick, S.; Gao, D.; Wang, A.~J.; Yan, R.;
  and Shou, M.~Z. 2023{\natexlab{b}}.
\newblock Univtg: Towards unified video-language temporal grounding.
\newblock In \emph{Proceedings of the IEEE/CVF International Conference on
  Computer Vision}, 2794--2804.

\bibitem[{Liu et~al.(2024)Liu, Li, Wu, and Lee}]{liu2024visual}
Liu, H.; Li, C.; Wu, Q.; and Lee, Y.~J. 2024.
\newblock Visual instruction tuning.
\newblock \emph{Advances in neural information processing systems}, 36.

\bibitem[{Luo et~al.(2023{\natexlab{a}})Luo, Huang, Gong, Jin, and
  Liu}]{luo2023towards}
Luo, D.; Huang, J.; Gong, S.; Jin, H.; and Liu, Y. 2023{\natexlab{a}}.
\newblock Towards generalisable video moment retrieval: Visual-dynamic
  injection to image-text pre-training.
\newblock In \emph{Proceedings of the IEEE/CVF Conference on Computer Vision
  and Pattern Recognition}, 23045--23055.

\bibitem[{Luo et~al.(2023{\natexlab{b}})Luo, Zhao, Yang, Dong, Qiu, Lu, Wang,
  and Wei}]{luo2023valley}
Luo, R.; Zhao, Z.; Yang, M.; Dong, J.; Qiu, M.; Lu, P.; Wang, T.; and Wei, Z.
  2023{\natexlab{b}}.
\newblock Valley: Video assistant with large language model enhanced ability.
\newblock \emph{arXiv preprint arXiv:2306.07207}.

\bibitem[{Maaz et~al.(2023)Maaz, Rasheed, Khan, and Khan}]{maaz2023video}
Maaz, M.; Rasheed, H.; Khan, S.; and Khan, F.~S. 2023.
\newblock Video-chatgpt: Towards detailed video understanding via large vision
  and language models.
\newblock \emph{arXiv preprint arXiv:2306.05424}.

\bibitem[{Moreno-Torres et~al.(2012)Moreno-Torres, Raeder,
  Alaiz-Rodr{\'\i}guez, Chawla, and Herrera}]{moreno2012unifying}
Moreno-Torres, J.~G.; Raeder, T.; Alaiz-Rodr{\'\i}guez, R.; Chawla, N.~V.; and
  Herrera, F. 2012.
\newblock A unifying view on dataset shift in classification.
\newblock \emph{Pattern recognition}, 45(1): 521--530.

\bibitem[{Oncescu et~al.(2021)Oncescu, Henriques, Liu, Zisserman, and
  Albanie}]{oncescu2021queryd}
Oncescu, A.-M.; Henriques, J.~F.; Liu, Y.; Zisserman, A.; and Albanie, S. 2021.
\newblock Queryd: A video dataset with high-quality text and audio narrations.
\newblock In \emph{ICASSP 2021-2021 IEEE International Conference on Acoustics,
  Speech and Signal Processing (ICASSP)}, 2265--2269. IEEE.

\bibitem[{Puigcerver et~al.(2023)Puigcerver, Riquelme, Mustafa, and
  Houlsby}]{puigcerver2023sparse}
Puigcerver, J.; Riquelme, C.; Mustafa, B.; and Houlsby, N. 2023.
\newblock From sparse to soft mixtures of experts.
\newblock \emph{arXiv preprint arXiv:2308.00951}.

\bibitem[{Qian et~al.(2024)Qian, Li, Wu, Ye, Fei, Chua, Zhuang, and
  Tang}]{qian2024momentor}
Qian, L.; Li, J.; Wu, Y.; Ye, Y.; Fei, H.; Chua, T.-S.; Zhuang, Y.; and Tang,
  S. 2024.
\newblock Momentor: Advancing Video Large Language Model with Fine-Grained
  Temporal Reasoning.
\newblock arXiv:2402.11435.

\bibitem[{Ren et~al.(2023)Ren, Yao, Li, Sun, and Hou}]{ren2023timechat}
Ren, S.; Yao, L.; Li, S.; Sun, X.; and Hou, L. 2023.
\newblock TimeChat: A Time-sensitive Multimodal Large Language Model for Long
  Video Understanding.
\newblock \emph{arXiv preprint arXiv:2312.02051}.

\bibitem[{Song et~al.(2024{\natexlab{a}})Song, Chai, Wang, Zhang, Zhou, Wu,
  Chi, Guo, Ye, Zhang et~al.}]{song2024moviechat}
Song, E.; Chai, W.; Wang, G.; Zhang, Y.; Zhou, H.; Wu, F.; Chi, H.; Guo, X.;
  Ye, T.; Zhang, Y.; et~al. 2024{\natexlab{a}}.
\newblock Moviechat: From dense token to sparse memory for long video
  understanding.
\newblock In \emph{Proceedings of the IEEE/CVF Conference on Computer Vision
  and Pattern Recognition}, 18221--18232.

\bibitem[{Song et~al.(2024{\natexlab{b}})Song, Chai, Ye, Hwang, Li, and
  Wang}]{song2024moviechat+}
Song, E.; Chai, W.; Ye, T.; Hwang, J.-N.; Li, X.; and Wang, G.
  2024{\natexlab{b}}.
\newblock MovieChat+: Question-aware Sparse Memory for Long Video Question
  Answering.
\newblock \emph{arXiv preprint arXiv:2404.17176}.

\bibitem[{Song et~al.(2015)Song, Vallmitjana, Stent, and
  Jaimes}]{song2015tvsum}
Song, Y.; Vallmitjana, J.; Stent, A.; and Jaimes, A. 2015.
\newblock Tvsum: Summarizing web videos using titles.
\newblock In \emph{Proceedings of the IEEE conference on computer vision and
  pattern recognition}, 5179--5187.

\bibitem[{Sun et~al.(2023)Sun, Fang, Wu, Wang, and Cao}]{sun2023eva}
Sun, Q.; Fang, Y.; Wu, L.; Wang, X.; and Cao, Y. 2023.
\newblock Eva-clip: Improved training techniques for clip at scale.
\newblock \emph{arXiv preprint arXiv:2303.15389}.

\bibitem[{Tang et~al.(2019)Tang, Ding, Rao, Zheng, Zhang, Zhao, Lu, and
  Zhou}]{tang2019coin}
Tang, Y.; Ding, D.; Rao, Y.; Zheng, Y.; Zhang, D.; Zhao, L.; Lu, J.; and Zhou,
  J. 2019.
\newblock Coin: A large-scale dataset for comprehensive instructional video
  analysis.
\newblock In \emph{Proceedings of the IEEE/CVF Conference on Computer Vision
  and Pattern Recognition}, 1207--1216.

\bibitem[{Tong et~al.(2022)Tong, Song, Wang, and Wang}]{tong2022videomae}
Tong, Z.; Song, Y.; Wang, J.; and Wang, L. 2022.
\newblock Videomae: Masked autoencoders are data-efficient learners for
  self-supervised video pre-training.
\newblock \emph{Advances in neural information processing systems}, 35:
  10078--10093.

\bibitem[{Touvron et~al.(2023)Touvron, Lavril, Izacard, Martinet, Lachaux,
  Lacroix, Rozi{\`e}re, Goyal, Hambro, Azhar et~al.}]{touvron2023llama}
Touvron, H.; Lavril, T.; Izacard, G.; Martinet, X.; Lachaux, M.-A.; Lacroix,
  T.; Rozi{\`e}re, B.; Goyal, N.; Hambro, E.; Azhar, F.; et~al. 2023.
\newblock Llama: Open and efficient foundation language models.
\newblock \emph{arXiv preprint arXiv:2302.13971}.

\bibitem[{Vedantam, Lawrence~Zitnick, and Parikh(2015)}]{vedantam2015cider}
Vedantam, R.; Lawrence~Zitnick, C.; and Parikh, D. 2015.
\newblock Cider: Consensus-based image description evaluation.
\newblock In \emph{Proceedings of the IEEE conference on computer vision and
  pattern recognition}, 4566--4575.

\bibitem[{Wang et~al.(2024{\natexlab{a}})Wang, Li, Li, Yu, He, Chen, Pei,
  Zheng, Xu, Wang et~al.}]{wang2024internvideo2}
Wang, Y.; Li, K.; Li, X.; Yu, J.; He, Y.; Chen, G.; Pei, B.; Zheng, R.; Xu, J.;
  Wang, Z.; et~al. 2024{\natexlab{a}}.
\newblock Internvideo2: Scaling video foundation models for multimodal video
  understanding.
\newblock \emph{arXiv preprint arXiv:2403.15377}.

\bibitem[{Wang et~al.(2022)Wang, Li, Li, He, Huang, Zhao, Zhang, Xu, Liu, Wang
  et~al.}]{wang2022internvideo}
Wang, Y.; Li, K.; Li, Y.; He, Y.; Huang, B.; Zhao, Z.; Zhang, H.; Xu, J.; Liu,
  Y.; Wang, Z.; et~al. 2022.
\newblock Internvideo: General video foundation models via generative and
  discriminative learning.
\newblock \emph{arXiv preprint arXiv:2212.03191}.

\bibitem[{Wang et~al.(2024{\natexlab{b}})Wang, Meng, Liang, Wang, Liu, and
  Zhao}]{wang2024hawkeye}
Wang, Y.; Meng, X.; Liang, J.; Wang, Y.; Liu, Q.; and Zhao, D.
  2024{\natexlab{b}}.
\newblock HawkEye: Training Video-Text LLMs for Grounding Text in Videos.
\newblock \emph{arXiv preprint arXiv:2403.10228}.

\bibitem[{Wang et~al.(2024{\natexlab{c}})Wang, Wang, Wu, Liang, Zhao, Liu, and
  Zheng}]{wang2024efficient}
Wang, Y.; Wang, Y.; Wu, P.; Liang, J.; Zhao, D.; Liu, Y.; and Zheng, Z.
  2024{\natexlab{c}}.
\newblock Efficient Temporal Extrapolation of Multimodal Large Language Models
  with Temporal Grounding Bridge.
\newblock In \emph{Proceedings of the 2024 Conference on Empirical Methods in
  Natural Language Processing}, 9972--9987.

\bibitem[{Wu et~al.(2024)Wu, Hu, Sun, Zhou, Zhu, Rao, Schiele, and
  Yang}]{wu2024number}
Wu, Y.; Hu, X.; Sun, Y.; Zhou, Y.; Zhu, W.; Rao, F.; Schiele, B.; and Yang, X.
  2024.
\newblock Number it: Temporal Grounding Videos like Flipping Manga.
\newblock \emph{arXiv preprint arXiv:2411.10332}.

\bibitem[{Xiao et~al.(2023)Xiao, Luo, Liu, Ma, Bian, Ji, Yang, and
  Li}]{xiao2023bridging}
Xiao, Y.; Luo, Z.; Liu, Y.; Ma, Y.; Bian, H.; Ji, Y.; Yang, Y.; and Li, X.
  2023.
\newblock Bridging the Gap: A Unified Video Comprehension Framework for Moment
  Retrieval and Highlight Detection.
\newblock \emph{arXiv preprint arXiv:2311.16464}.

\bibitem[{Xu et~al.(2021)Xu, Ghosh, Huang, Okhonko, Aghajanyan, Metze,
  Zettlemoyer, and Feichtenhofer}]{xu2021videoclip}
Xu, H.; Ghosh, G.; Huang, P.-Y.; Okhonko, D.; Aghajanyan, A.; Metze, F.;
  Zettlemoyer, L.; and Feichtenhofer, C. 2021.
\newblock Videoclip: Contrastive pre-training for zero-shot video-text
  understanding.
\newblock \emph{arXiv preprint arXiv:2109.14084}.

\bibitem[{Yang et~al.(2023)Yang, Nagrani, Seo, Miech, Pont-Tuset, Laptev,
  Sivic, and Schmid}]{yang2023vid2seq}
Yang, A.; Nagrani, A.; Seo, P.~H.; Miech, A.; Pont-Tuset, J.; Laptev, I.;
  Sivic, J.; and Schmid, C. 2023.
\newblock Vid2seq: Large-scale pretraining of a visual language model for dense
  video captioning.
\newblock In \emph{Proceedings of the IEEE/CVF Conference on Computer Vision
  and Pattern Recognition}, 10714--10726.

\bibitem[{Zala et~al.(2023{\natexlab{a}})Zala, Cho, Kottur, Chen, Oguz, Mehdad,
  and Bansal}]{zala2023hierarchical}
Zala, A.; Cho, J.; Kottur, S.; Chen, X.; Oguz, B.; Mehdad, Y.; and Bansal, M.
  2023{\natexlab{a}}.
\newblock Hierarchical video-moment retrieval and step-captioning.
\newblock In \emph{Proceedings of the IEEE/CVF Conference on Computer Vision
  and Pattern Recognition}, 23056--23065.

\bibitem[{Zala et~al.(2023{\natexlab{b}})Zala, Cho, Kottur, Chen, Oğuz,
  Mehdad, and Bansal}]{Zala2023HiREST}
Zala, A.; Cho, J.; Kottur, S.; Chen, X.; Oğuz, B.; Mehdad, Y.; and Bansal, M.
  2023{\natexlab{b}}.
\newblock Hierarchical Video-Moment Retrieval and Step-Captioning.
\newblock In \emph{CVPR}.

\bibitem[{Zellers et~al.(2021)Zellers, Lu, Hessel, Yu, Park, Cao, Farhadi, and
  Choi}]{zellersluhessel2021merlot}
Zellers, R.; Lu, X.; Hessel, J.; Yu, Y.; Park, J.~S.; Cao, J.; Farhadi, A.; and
  Choi, Y. 2021.
\newblock MERLOT: Multimodal Neural Script Knowledge Models.
\newblock In \emph{Advances in Neural Information Processing Systems 34}.

\bibitem[{Zhang, Li, and Bing(2023)}]{zhang2023video}
Zhang, H.; Li, X.; and Bing, L. 2023.
\newblock Video-llama: An instruction-tuned audio-visual language model for
  video understanding.
\newblock \emph{arXiv preprint arXiv:2306.02858}.

\bibitem[{Zhao et~al.(2024)Zhao, Gundavarapu, Yuan, Zhou, Yan, Sun, Friedman,
  Qian, Weyand, Zhao et~al.}]{zhao2024videoprism}
Zhao, L.; Gundavarapu, N.~B.; Yuan, L.; Zhou, H.; Yan, S.; Sun, J.~J.;
  Friedman, L.; Qian, R.; Weyand, T.; Zhao, Y.; et~al. 2024.
\newblock VideoPrism: A Foundational Visual Encoder for Video Understanding.
\newblock \emph{arXiv preprint arXiv:2402.13217}.

\bibitem[{Zhou, Xu, and Corso(2018)}]{zhou2018towards}
Zhou, L.; Xu, C.; and Corso, J. 2018.
\newblock Towards automatic learning of procedures from web instructional
  videos.
\newblock In \emph{Proceedings of the AAAI Conference on Artificial
  Intelligence}, volume~32.

\bibitem[{Zhu et~al.(2023)Zhu, Chen, Shen, Li, and Elhoseiny}]{zhu2023minigpt}
Zhu, D.; Chen, J.; Shen, X.; Li, X.; and Elhoseiny, M. 2023.
\newblock Minigpt-4: Enhancing vision-language understanding with advanced
  large language models.
\newblock \emph{arXiv preprint arXiv:2304.10592}.

\end{thebibliography}

{
    \parskip=0em
    \renewcommand{\contentsname}{Contents of Appendix}
    \tableofcontents
    \addtocontents{toc}{\protect\setcounter{tocdepth}{3}}
}

\begin{table}
    \centering
        \begin{tabular}{l c c c c c c c c c}
            \toprule
            \multirow{2}{*}{Model} & \multicolumn{3}{c}{Youcook2}                         \\
            \cmidrule(lr){2-4}
                                       & SODA\_c                                                          & CIDEr                                                         & F1 Score                                                          \\
            \midrule
            \textit{Task-Specific Models} \\
            Vid2Seq & \texttransparent{0.5}{7.9} & \texttransparent{0.5}{ 47.1} & \texttransparent{0.5}{ 27.3}\\
            Vid2Seq (Visual Only) & \texttransparent{0.5}{5.7} & \texttransparent{0.5}{ 25.3} & \texttransparent{0.5}{ 23.5} \\
            \midrule
            \textit{Generalist Models} \\
            TimeChat & 3.4 & 11.0 & 19.5 \\
            VTG-LLM ($\tau = 1.0$) & 3.6 & 13.4 & 20.6 \\
            VTG-LLM ($\tau = 0.1$) & \textbf{3.9} & \textbf{13.7} & \textbf{21.0} \\
            \bottomrule
        \end{tabular}%
    \caption{
        \textbf{Fine-tuned performance of algorithms on Youcook2 dataset.}
        We fine-tune the algorithms on Youcook2 dataset with a batch size of 16. We use transparent color for traditional non-generalist models and task-specific models. Vid2Seq~\citep{yang2023vid2seq} fully training the LLM models and using 1B video data for pretraining.
        \looseness=-1
    }
    \label{tab:ft-vtg-youcook}
\end{table}

\begin{table}[!t]
    \centering
        \begin{tabular}{l c c c c c c c c c}
            \toprule
            \multirow{2}{*}{Model} & \multicolumn{2}{c}{Charades-STA}                         \\
            \cmidrule(lr){2-3}
                                       & $\text{R@1}_{\text{(IOU=0.5)}}$                                                         & $\text{R@1}_{\text{(IOU=0.7)}}$                                                         \\
            \midrule
            \textit{Non-Generative Models} \\
            InternVideo2-6B &  \texttransparent{0.5}{70.0} & \texttransparent{0.5}{49.0} \\
            VDI  & \texttransparent{0.5}{52.3} & \texttransparent{0.5}{31.4} \\
            Moment-DETR & \texttransparent{0.5}{55.7} & \texttransparent{0.5}{34.2} \\
            \midrule
            \textit{Task-Specific Models} \\
            HawkEye & \texttransparent{0.5}{58.3 } & \texttransparent{0.5}{28.8 }\\
            \midrule
            \textit{Generalist Models} \\
            TimeChat & 46.7 & 23.7 \\
            VTG-LLM ($\tau = 1.0$) & 57.2 & 33.4 \\
            VTG-LLM ($\tau = 0.1$) & \textbf{57.8} & \textbf{33.9} \\
            \bottomrule
        \end{tabular}%
    \caption{
        \textbf{Fine-tuned performance of algorithms on Charades-STA dataset.}
        We fine-tune the Algorithm on Charades-STA datasets with a batch size of 16. We use transparent color for traditional non-generalist models and task-specific models. We choose InternVideo2-6B~\citep{wang2024internvideo2}, VDI~\citep{luo2023towards}, and Moment-DETR~\citep{lei2021detecting} as examples of non-generative models.
        \looseness=-1
    }
    \label{tab:ft-vtg-charades}
\end{table}

\begin{table}[!t]
    \centering
        \begin{tabular}{l c c c c c c c c c}
            \toprule
            \multirow{2}{*}{Model} & \multicolumn{2}{c}{QVHighlights}                         \\
            \cmidrule(lr){2-3}
                                       & $\text{R@1}_{\text{(IOU=0.5)}}$                                                         & $\text{R@1}_{\text{(IOU=0.7)}}$                                                         \\
            \midrule
            \textit{Non-Generative Models} \\
            Moment-DETR &  \texttransparent{0.5}{37.4} & \texttransparent{0.5}{60.2}\\
            HL-CLIP &  \texttransparent{0.5}{41.9} & \texttransparent{0.5}{70.6} \\
            \midrule
            \textit{Generalist Models} \\
            TimeChat & 21.7 & 37.9 \\
            VTG-LLM ($\tau = 1.0$) & \textbf{24.1} & \textbf{41.3} \\
            VTG-LLM ($\tau = 0.1$) & 23.0 & 40.8 \\
            \bottomrule
        \end{tabular}%
    \caption{
        \textbf{Fine-tuned performance of algorithms on QVHighlights datasets.}
        We fine-tune the Algorithm on QVHighlights datasets with a batch size of 16. We use transparent color for traditional non-generalist models and task-specific models. We choose Moment-DETR~\citep{lei2021detecting} and HL-CLIP~\citep{han2024unleash} as examples of non-generative models.
        \looseness=-1
    }
    \label{tab:ft-vtg-qvhighlights}
\end{table}

\begin{table*}
    \centering
        \begin{tabular}{l c c c c c c c c c}
            \toprule
            \multirow{2}{*}{Model} & \multicolumn{3}{c}{Youcook2}                         & \multicolumn{2}{c}{Charades-STA}                      & \multicolumn{2}{c}{QVHighlights}                                                                                                                                                                                                                                    \\
            \cmidrule(lr){2-4} \cmidrule(lr){5-6} \cmidrule(lr){7-8}
                                       & SODA\_c                                                          & CIDEr                                                         & F1 Score                                                          & $\text{R@1}_{\text{(IOU=0.5)}}$                                                         & $\text{R@1}_{\text{(IOU=0.7)}}$                                                                   & mAP & HIT@1                                                                 \\
            \midrule
            \textit{Temperature $\tau$} \\
            $\tau = 1.0$ & 1.5 & 5.0 & 17.5 & 33.8 & 15.7 & 16.5 & 33.5\\
            $\tau = 0.1$ & 1.6 & 5.4 & 18.4 & 36.3 & 16.6 & 16.2 & 30.7 \\
            \midrule
            \textit{Test-time interpolation} \\
            w/ interpolation & 1.5 & 5.0 & 17.5 & 33.8 & 15.7 & 16.5 & 33.5\\
            w/o interpolation & 1.5 & 5.0 & 17.0 & 33.7 & 15.5 & 16.3 & 32.4 \\
            \bottomrule
        \end{tabular}%
    \caption{
        \textbf{Additional ablation studies of VTG-LLM.}
        We conduct additional ablation studies in Table~\ref{tab:additional-ablation}. 
        \looseness=-1
    }
    \label{tab:additional-ablation}
\end{table*}

\begin{table}
    \centering
        \begin{tabular}{l c c c c c c c c c}
            \toprule
            \multirow{2}{*}{Model} & \multicolumn{3}{c}{Youcook2}                         \\
            \cmidrule(lr){2-4}
                                       & SODA\_c                                                          & CIDEr                                                         & F1 Score                                                          \\
            \midrule
            Momentor & 2.3 & 14.9 \\
            TimeChat & 4.7 & 19.0 & \textbf{36.9} \\
            VTG-LLM ($\tau = 1.0$) & 4.7 & 18.2 & 34.0 \\
            VTG-LLM ($\tau = 0.1$) & \textbf{5.1} & \textbf{20.7} & 34.8 \\
            \bottomrule
        \end{tabular}%
    \caption{
        \textbf{Peformance of algorithms on ActivityNet Captions dataset.}
    }
    \label{tab:anet-captions}
\end{table}

\section{Details on construction of VTG-IT}

We re-annotate the yttemporal part of Time-IT and use the original annotation of other parts. In detail,
\begin{itemize}
    \item VTG-IT-DVC and VTG-IT-VS: We utilize the Gemini-1.5 Pro to directly generate these two datasets. The prompts are provided in Table~\ref{tab:prompts}.
    \item VTG-IT-MR: We employ the VTG-IT-DVC to generate the VTG-IT-MR by utilizing descriptions as queries and timestamps as answers.
    \item For a random subset of VTG-IT-VS, for each description, we segment the videos and use a CLIP model to compute a similarity score between segments and description. Scores are normalized to a range of [1, 5]. We use descriptions as queries, and the scores and timestamps as answers. Low-score segments are filtered out.
\end{itemize}

\newcolumntype{Y}{>{\arraybackslash}m{0.1\textwidth}}
\newcolumntype{Z}{>{\arraybackslash}m{0.7\textwidth}}
\begin{table*}[!t]
    \centering
    \begin{tabularx}{\textwidth}{
    Y Z
    }
    \toprule
    Task & Prompt \\
    \midrule
    DVC & Please locate a series of events in the video, output the start and end timestamps of each event, and describe each event in sentences. The output format of each predicted event should be as follows: "Start - End seconds, event description". A specific example is: "90 - 102 seconds, spreading butter on two slices of white bread". Please take care of the output format. \\
    \midrule
    VS & Please find the highlight frames in the video and mark the timestamps of the highlight frames and a significance score of 1-5. The output format should be as follows: "Second of highlight frames, significance score, description". A specific example is: "At 82 second, significance score: 5, the detailed demonstration of how to use the round brush to create volume and waves, which is the core technique for achieving the final look." Please take care of the output format. \\
    \bottomrule
    \end{tabularx}
    \caption{\textbf{Prompts for constructing VTG-IT.}}
    \label{tab:prompts}
\end{table*}

\section{Experiments}

\subsection{Detailed Experiment Settings}

In this section, we present the detailed experimental settings used in our study. We employ ViT-G/14 from EVA-CLIP~\citep{sun2023eva} and Qformer from InstructBLIP~\citep{dai2024instructblip} as the visual backbone. The number of queries for Qformer is set to 32. The vision projection layer and the weights of the sequence embedding are initialized using the Video-LLaMA checkpoint~\citep{zhang2023video}. To address the mismatched size of the sequence embedding, we use interpolation to extend the size from 32 to 96. The time-embedding (8192) is initialized with zeros. We set the number of slots to 256.

For the language model backbone, we utilize LLaMA-2-7B~\citep{touvron2023llama}. The LoRA training is conducted on "q\_proj", "k\_proj", "v\_proj", and "o\_proj". Special time tokens are added to the LLaMA tokenizer, and we set the prediction head of LLM and the weights of token embedding to be trainable. The maximum text length is set to 2048.

The models are trained on the \dataset-120k and a randomly sampled subset (97k) from the Valley dataset~\citep{luo2023valley}. We sample 96 frames for each video. Training is carried out using 16 ATN 910B GPUs with a batch size of 64 and takes approximately 40 hours to complete. The training employs DDP and requires about 30GB of GPU storage for each GPU. Fine-tuning also uses 16 GPUs and sets the batch size to 16. 
We fine-tune the \alg model for 10 epochs on Charades dataset, 16 epochs for Youcook2 dataset, and 20 epochs for QVHighlights dataset.
We set the initial learning rate to 3e-5 and train the models for 10 epochs by default. The weight decay is set to 0.05, and the warm-up steps use $0.6E$, where $E$ represents the number of iterations for each epoch.

\subsection{Additional Experiment Results}


\paragraph{Fine-tuned model performance of VTG-LLM.} In Tables~\ref{tab:ft-vtg-youcook}, ~\ref{tab:ft-vtg-charades}, and~\ref{tab:ft-vtg-qvhighlights}, we fine-tune the VTG-LLM models on Youcook2, Charades-STA, and QVHighlights datasets. From the results, we make the following observations: (1) The performance of VTG-LLM is significantly better than other generalist models across all three tasks; (2) The performance of VTG-LLM on the moment retrieval task (Charades-STA) is comparable to task-specific models like HawkEye, as well as non-generative models like VDI and Moment-DETR.
(3) The task-specific model Vid2Seq~\citep{yang2023vid2seq} achieves superior performance on the Youcook2 dataset by fully training the LLM models and using 1B video data for pretraining. In contrast, we only fine-tune the LLM using LoRA and employ 120K instruction tuning data.

\paragraph{Ablation studies on decoding temperature.} In Table~\ref{tab:additional-ablation}, we present the performance of VTG-LLM when employing various decoding temperatures. The results indicate that decreasing the temperature considerably enhances the performance for VTG tasks.

\paragraph{Ablation studies on test-time interpolation.} In Table~\ref{tab:additional-ablation}, we present the performance of VTG-LLM w/o test-time interpolation. Results show that test-time interpolation boost the performance of VTG-LLM on long video datasets like Youcook2.

\begin{figure*}[!t]
    \centering
    \includegraphics[width=.75\textwidth]{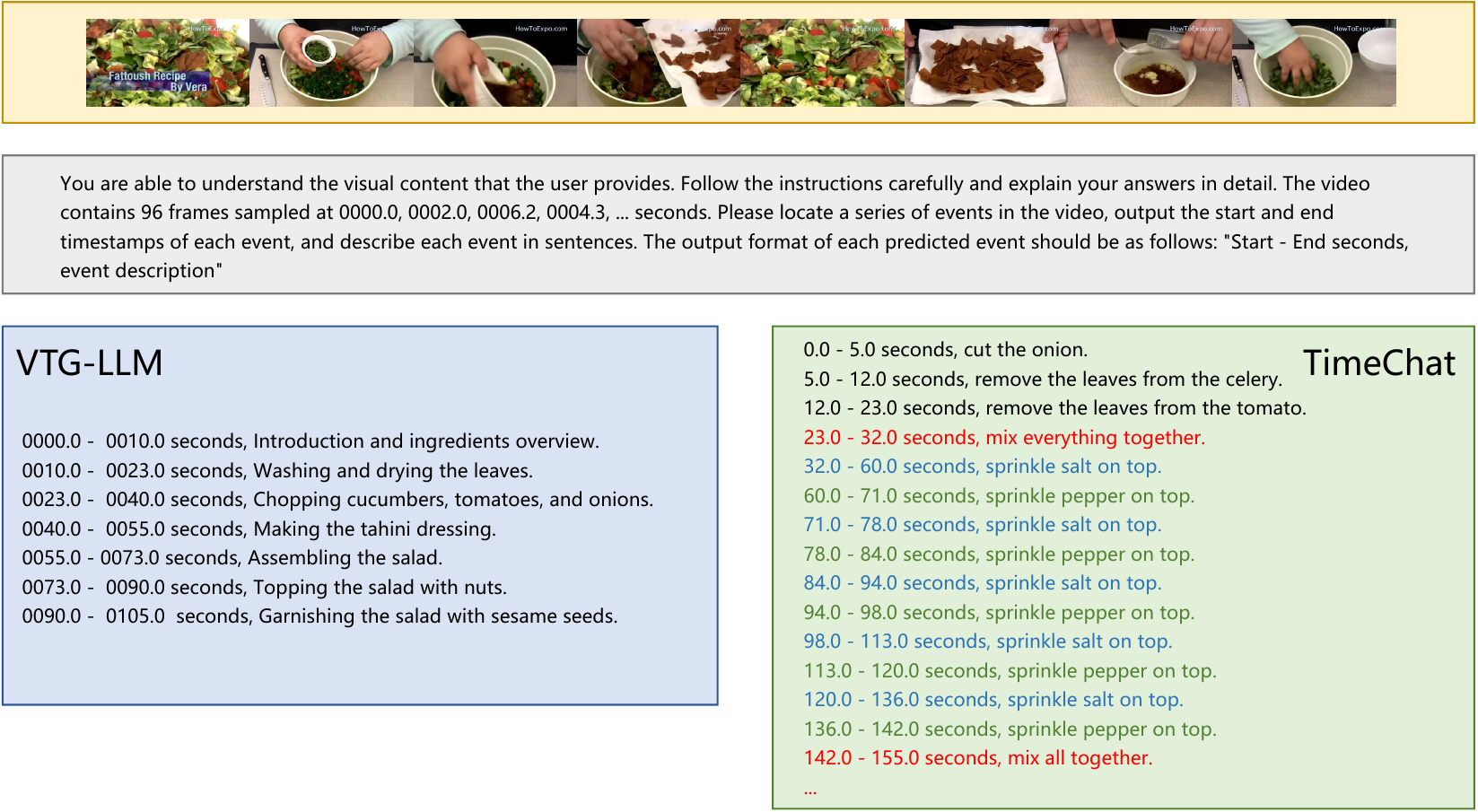}
    \caption{Quality analysis.}
    \label{fig:quality-analysis}
\end{figure*}

\paragraph{Quality analysis} In Figure~\ref{fig:quality-analysis}, we compare the outputs of VTG-LLM and TimeChat in a zero-shot manner on the Youcook2 dataset. We discovered that VTG-LLM generates more detailed and high-quality captions than TimeChat.

\begin{figure}[!t]
    \centering
    \includegraphics[width=.5\textwidth]{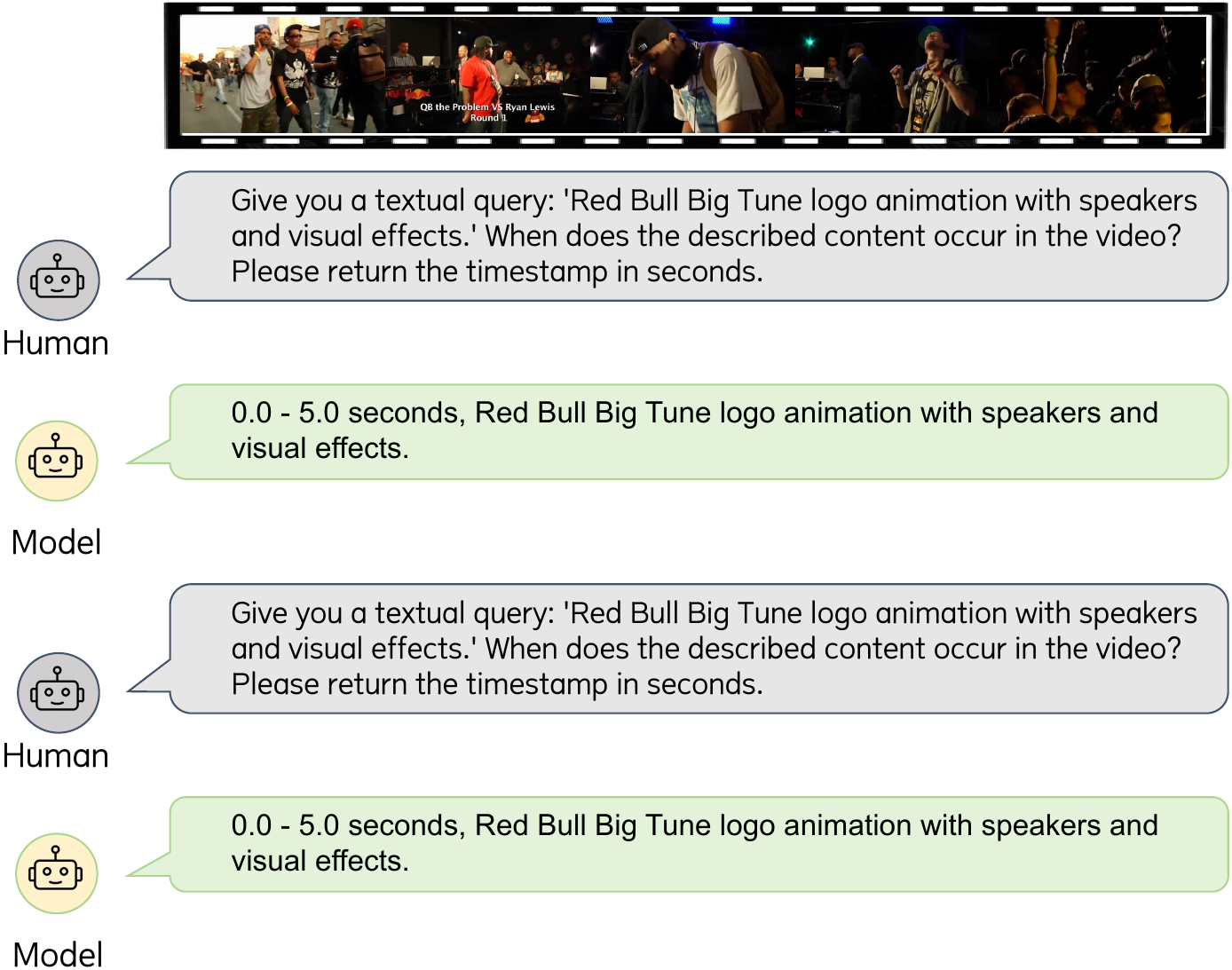}
    \caption{Annotation example of moment retrieval task.}
    \label{fig:example-mr}
\end{figure}

\begin{figure}[!t]
    \centering
    \includegraphics[width=.5\textwidth]{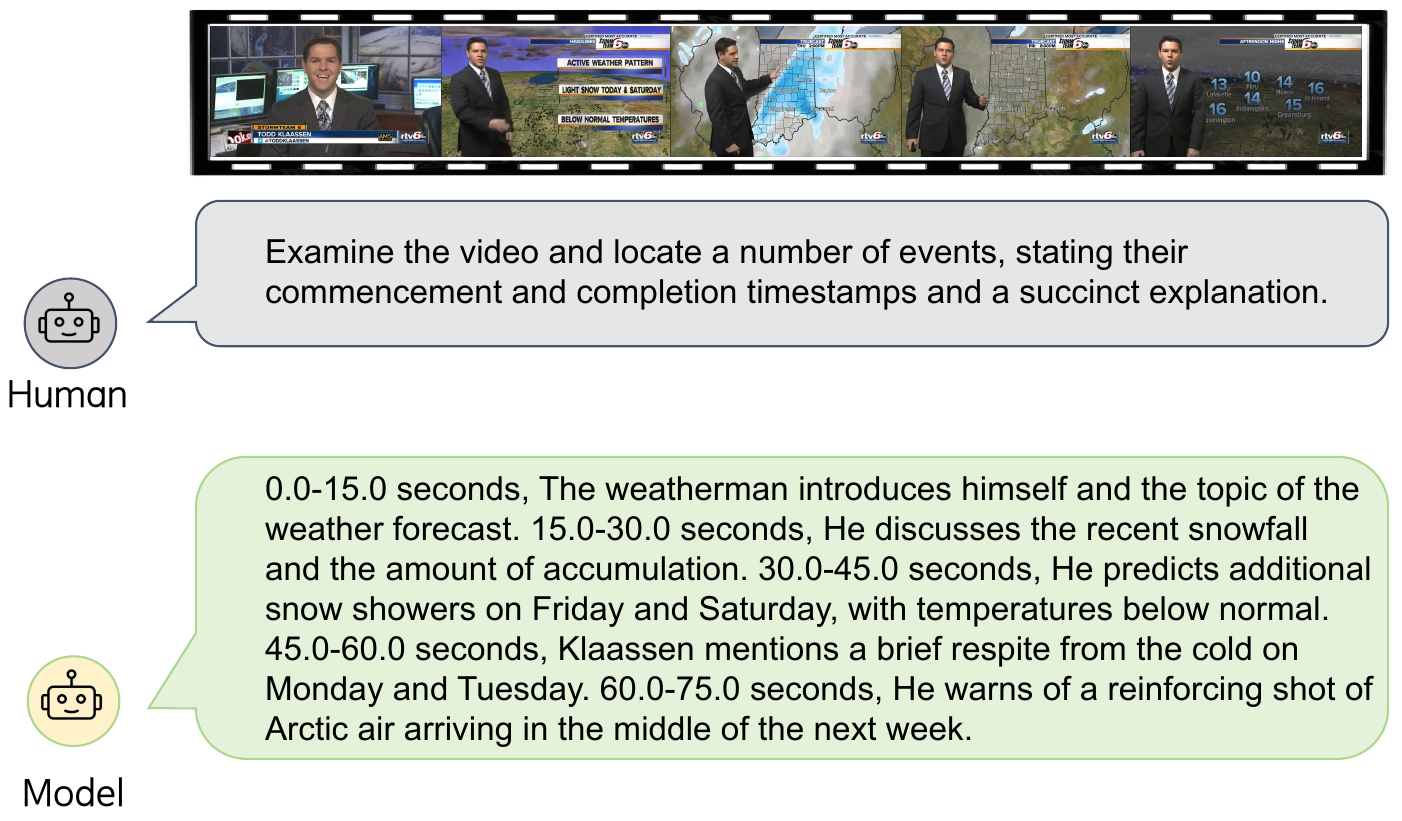}
    \caption{Annotation example of dense video captioning  task.}
    \label{fig:example-dvc}
\end{figure}

\begin{figure}[!t]
    \centering
    \includegraphics[width=.5\textwidth]{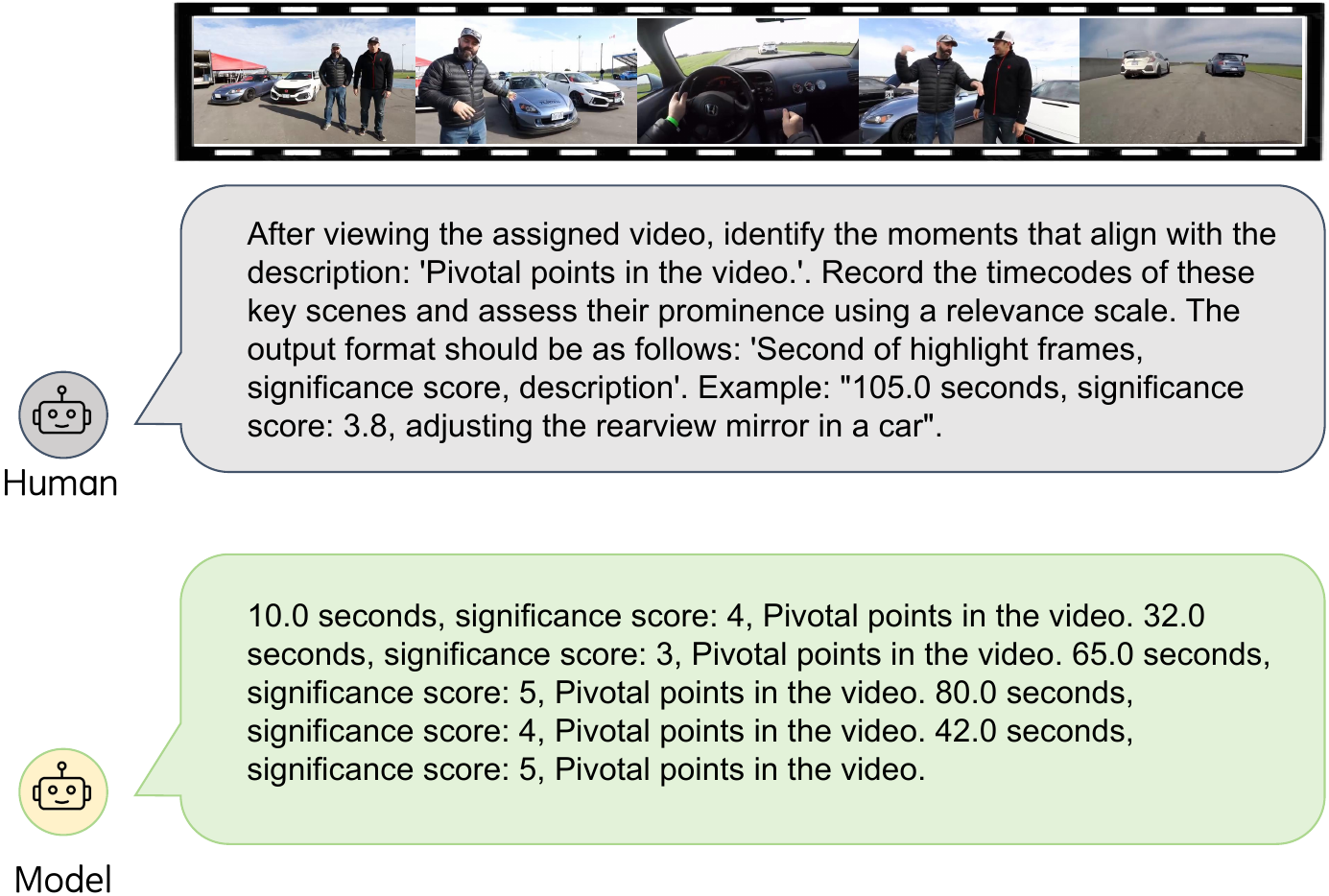}
    \caption{Annotation example of video summarization  task.}
    \label{fig:example-vs}
\end{figure}



\section{Additional Case Studies}
\label{sec:case-study}

\begin{figure}[!t]
    \centering
    \includegraphics[width=.5\textwidth]{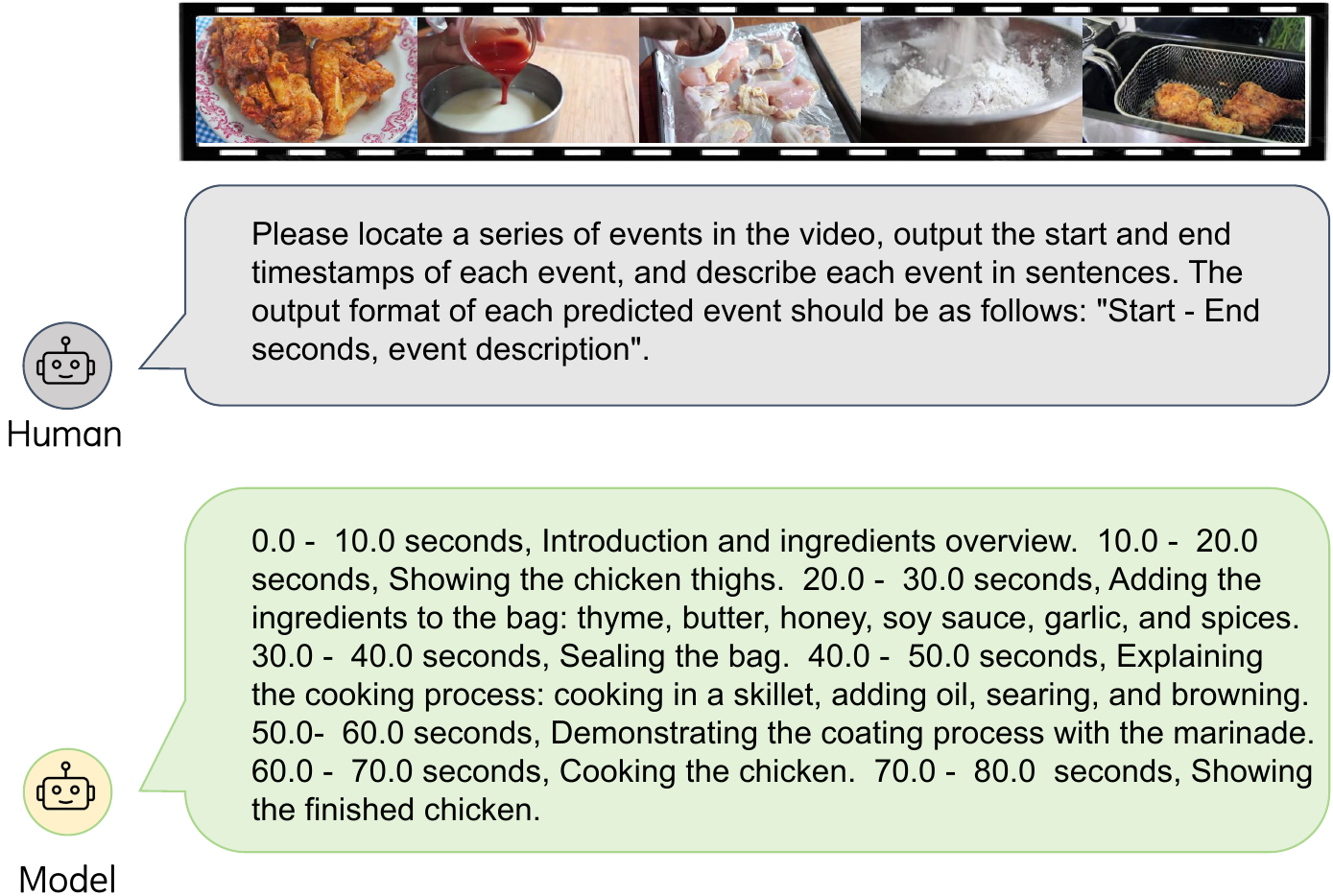}
    \caption{Zero-shot case study of dense video captioning  task.}
    \label{fig:case-dvc}
\end{figure}

We present annotation examples (Figure~\ref{fig:example-mr}, ~\ref{fig:example-vs} and ~\ref{fig:example-dvc}) of our new annotation. 
Moreover, we show the additional zero-shot case studies on dense video captioning  (Figure~\ref{fig:case-dvc}) task.

\end{document}